\documentclass[twoside,twocolumn]{article}

\newcommand{\Appendix}[1]{the full version for}

\renewcommand{\v}{\bm{v}}

\newcommand{\x}{\bm{x}}

\newcommand{\z}{\bm{z}}

\newcommand{\C}{\bm{C}}

\newcommand{\E}{\bm{E}}

\newcommand{\I}{\bm{I}}

\renewcommand{\Re}{\mathbb{R}}

\newcommand{\U}{\bm{U}}

\newcommand{\X}{\bm{X}}

\newcommand{\Z}{\bm{Z}}

\renewcommand{\mathbf}{\boldsymbol}
\usepackage{fitee}

\usepackage[utf8]{inputenc} 
\usepackage[T1]{fontenc}    
\usepackage{url}            
\usepackage{booktabs}       
\usepackage{amsfonts}       
\usepackage{nicefrac}       
\usepackage{microtype}      
\usepackage[dvipsnames]{xcolor}         
\usepackage{bbding}
\usepackage{pifont}

\usepackage{bm}
\usepackage[normalem]{ulem}
\usepackage{times,epsfig,graphics,graphicx,subfigure,wrapfig,float,booktabs,xcolor,multirow,multicol,array,color,ifthen,tabu,colortbl,url,xparse,mathtools,patchcmd,enumitem,amssymb,xspace,nicefrac,microtype,amsmath,amsfonts,amsthm}

\definecolor{applegreen}{rgb}{0.55, 0.71, 0.0}
\definecolor{cornflowerblue}{rgb}{0.39, 0.58, 0.93}

\usepackage[
    CJKbookmarks=true,
    bookmarksnumbered=true,
    bookmarksopen=true,
    colorlinks,
    linktocpage,
    linkcolor={blue!70!gray}, citecolor={blue!70!gray}, urlcolor={blue!70!gray}
]{hyperref}
\addto{\captionsenglish}{%
  
}

\begin{document}

\title{On the Principles of Parsimony and Self-Consistency\\for the Emergence of Intelligence}

\author[$\dagger$$\ddagger$1]{Yi Ma}%
\author[$\dagger$2]{Doris Tsao}
\author[$\dagger$3]{Heung-Yeung Shum}
\affil[1]{Electrical Engineering and Computer Science Department, University of California, Berkeley, CA 94720, USA}
\affil[2]{Department of Molecular \& Cell Biology and Howard Hughes Medical Institute, University of California, Berkeley, CA 94720, USA}
\affil[3]{International Digital Economy Academy, Shenzhen 518045, China}
\shortauthor{Ma, Tsao and Shum}	

\authmark{}



\corremailA{yima@eecs.berkeley.edu}
\corremailB{dortsao@berkeley.edu}
\corremailC{hshum@idea.edu.cn}
\emailmark{$\dagger$}	


\abstract{Ten years into the revival of deep networks and artificial intelligence, we propose a theoretical framework that sheds light on understanding deep networks within a bigger picture of intelligence in general. We introduce two fundamental principles, {\em Parsimony} and {\em Self-consistency}, which address two fundamental questions regarding intelligence: what to learn and how to learn, respectively. We believe the two principles serve as the cornerstone for the emergence of intelligence, artificial or natural. While they have rich classical roots, we argue that they can be stated anew in entirely measurable and computable ways. More specifically, the two principles lead to an effective and efficient computational framework, compressive closed-loop transcription, which unifies and explains the evolution of modern deep networks and most practices of artificial intelligence. While we mainly use visual data modeling as an example, we believe the two principles will unify understanding of broad families of autonomous intelligent systems and provide a framework for understanding the brain.}



\doi{10.1631/FITEE.1000000}	
\code{A}
\clc{TP}




\maketitle

\section{Context and Motivation}


For an autonomous intelligent agent to survive and function in a complex environment, it must efficiently and effectively learn models that reflect both its past experiences and the current environment being perceived. Such models are critical for gathering information, making decisions, and taking action. 
Generally referred to as world models, these models should be continuously improved based on how projections agree with new observations and outcomes. 
They should incorporate both knowledge from past experiences (e.g., recognizing familiar objects) and mechanisms for interpreting immediate sensory inputs (e.g., detecting and tracking moving objects).  Studies in neuroscience suggest that the brain's world model is highly {\em structured} anatomically  (e.g., modular brain areas and columnar organization) and functionally (e.g., sparse coding \citep{olshausen1996emergence} and subspace coding \citep{Chang-Cell-2017,Bao2020AMO}). Such a structured model is believed to be the key to the brain’s efficiency and effectiveness in perceiving, predicting, and making intelligent decisions \citep{Barlow1961-ce,Josselyn2020MemoryER}.

In contrast, in the past decade, progress in artificial intelligence has relied mainly on training ``tried-and-tested'' models with largely homogeneous structures, like deep neural networks \citep{lecun2015deep}, using a 
brute-force engineering approach.
While functional modularity may emerge from training, the learned feature representation remains largely hidden or latent inside and is difficult to interpret \citep{Zeiler-ECCV2014}. Currently, such expensive brute-force end-to-end training of black-box models has  resulted in ever-growing model size and high data/computation cost\footnote{With model sizes frequently going beyond billions or trillions of parameters, even Google seems to recently have started worrying about the carbon footprint of such practices  \citep{Google-Carbon-2022}!}, and is accompanied by many caveats in practice: the lack of richness in final learned representations due to neural collapse \citep{papyan2020prevalence}\footnote{This refers to the final representation for each class collapsing to a one-hot vector that carries no information about the input except its class label. Richer features might be learned inside the networks, but their structures are unclear and remain largely hidden.}; lack of stability in training due to mode collapse \citep{veegan};  lack of adaptiveness and susceptibility to catastrophic forgetting \citep{catastrophic}; and lack of robustness to deformations \citep{azulay2018deep,engstrom2017rotation} or adversarial attacks \citep{szegedy2013intriguing}. 

\begin{figure*}[t]
\centering
\includegraphics[width=0.90\textwidth]{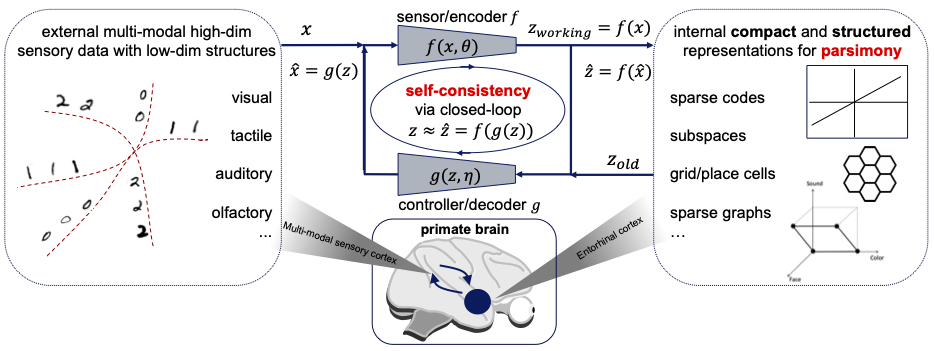}
\caption{\textbf{Overall framework} for a universal learning engine:  seeking a compact and structured model for sensory data via a  compressive closed-loop transcription: a (nonlinear) mapping $f(\cdot, \theta): \x \mapsto \z$ that maps high-dimensional sensory data with complicated low-dimensional structures to a compact structured representation. The model needs to be self-consistent, i.e., it can regenerate the original data via a map $g(\cdot, \eta): \z \mapsto \hat{\x}$ such that $f$ cannot distinguish despite its best effort. (Often, for simplicity, we omit the dependency of the mappings $f$ and $g$ on their parameters $\theta$ and $\eta$, respectively.)}
\vspace{-3mm}
\label{fig:framework}
\end{figure*}

\vspace{-2mm}
\paragraph{A principled and unifying approach?} We hypothesize that a fundamental reason why these problems arise in the current practice of deep networks and artificial intelligence is a lack of systematic and integrated understanding about the functional and organizational principles of intelligent systems. 

For instance, training discriminative models for classification and generative models for sampling or replaying has been largely separated in practice. Such models are typically open-loop systems that must be trained end-to-end via supervision or self-supervision. A principle long-learned in control theory is that such open-loop systems cannot automatically correct errors in prediction, and are unadaptive to changes in the environment. This had led to the introduction of ``closed-loop feedback'' to controlled systems so that a system can learn to correct its errors \citep{Wiener-1948,Mayr1970}. As we will argue in this paper, a similar lesson can be drawn here: once discriminative  and  generative models are combined to form a complete closed-loop system, learning can become autonomous (without exterior supervision), more efficient, stable, and adaptive. 

To understand any functional component that may be necessary for an intelligent system, such as a discriminative or a generative segment, we need to understand ntelligence from a more principled and unifying perspective. Therefore, in this paper, we introduce {two fundamental principles}: {\em Parsimony} and {\em Self-consistency}, which we believe govern the function and design of any intelligent system, artificial or natural. The two principles aim to answer the following two fundamental questions regarding learning, respectively:
\begin{enumerate}
    \item {\em What to learn:} what is the objective of  learning from data, and how can it be measured?
    \item {\em How to learn:} how can we achieve such an objective via efficient and effective computation?
\end{enumerate}

As we will see, answers to the first question fall naturally into the realm of Information/Coding Theory \citep{Shannon}, which studies how to accurately {\em quantify and measure} the information in the data and then seek {\em  the most compact} representations of the information.  Once the objective of learning is clear and set, answers to the second question fall naturally into the realm of Control/Game Theory \citep{Wiener-1948}, which provides a universally effective computational framework, i.e., a {\em closed-loop} feedback system, for achieving any measurable objective {\em consistently} (Figure \ref{fig:framework}).

The basic ideas behind each of the two principles proposed in this paper can find their roots in classic works. Artificial (deep) neural networks, since their earliest inception as ``perceptrons'' \citep{Rosenblatt1958ThePA}, were conceived to store and organize sensory information efficiently. Back propagation \citep{HenryKelley,Back-Prop} was later proposed as a mechanism for learning such models. Moreover, even before the inception of neural networks, Norbert Wiener had contemplated computational mechanisms for learning at a system level. In his famed book {\em Cybernetics} \citep{Wiener-1961}, he studied the possible roles of information compression for parsimony and feedback/games in a learning machine for consistency. 

But we are here to reunite and restate the two principles within the new context of data science and machine learning, as they help better explain and unify most modern instances and practices of artificial intelligence, particularly deep learning.\footnote{As we will see, besides integrating discriminative and generative models, they lead to a closed-loop framework that works uniformly in supervised, incremental or unsupervised settings, without suffering many of the problems of open-loop deep networks.} Different from previous efforts, our restatement of these principles will be entirely {\em measurable} and {\em computationally tractable} -- hence easily realizable by machines or in nature with limited resources. This paper aims to offer our overall position and perspective rather than to justify every claim technically. Nevertheless, we will provide references to related work where readers can find convincing theoretical and compelling empirical evidence. They are based on a coherent series of past and recent developments in the study of machine learning and data science by the authors and their students  \citep{ma2007segmentation,wright2008classification,chan2015pcanet,yu2020learning,chan2021redunet,Baek-CVPR2022,dai2021closedloop,tong2022incremental,Druv-Pai,Wright-Ma-2021}. 

\vspace{-2mm}
\paragraph{Organization.} In Section \ref{sec:two-principles}, we  use visual data modeling as a concrete  example to introduce the two principles and illustrate how they can be instantiated as computable objectives,  architectures, and systems. 
In Section \ref{sec:learning-engine}, we conjecture that they lead to a universal learning engine for broader perception and decision making tasks. Finally, in Section \ref{sec:broad-program}, we discuss several implications of the proposed principles and their connections to neuroscience, mathematics, and higher-level intelligence.

\section{Two Principles for Intelligence}
\label{sec:two-principles}
\vspace{-2mm}

In this section, we introduce and explain the two fundamental principles that can help answer the questions of {\it what to learn} and {\it how to learn} by an intelligent agent or system. 

\subsection{What to Learn: the Principle of Parsimony}
\label{sec:parsimony}

\begin{quote}
``{\em Entities should not be multiplied unnecessarily}.'' 
\vspace{-2mm}

$~$\hfill -- William of Ockham
\vspace{-2mm}
\end{quote}

\paragraph{The Principle of Parsimony:} {\em The objective of learning for an intelligent system is to identify low-dimensional structures in observations of the external world and reorganize them in the most compact and structured way.
} 


 There is a fundamental reason why intelligent systems need to embody this principle: intelligence would be impossible without it! If observations of the external world had no low-dimensional structures, nothing would be worth learning or memorizing. Nothing could be relied upon for good generalization or prediction, which rely on new observations following the same low-dimensional structures. Thus, this {\it principle} is not simply a {\it convenience} arising from the need for intelligent systems to be frugal with their resources, such as energy, space, time, matter, etc.

In some contexts, the above principle is also called {the Principle of Compression}. But Parsimony of intelligence is not about achieving the best possible compression, but about {\it obtaining compact and  structured representations via computationally efficient means}. There is no point for an intelligent system to try to compress data to the ultimate level of Kolmogorov complexity or Shannon information: they are not only intractable to compute (or even to approximate) but also result in completely unstructured representations. For instance, representing data with the minimum description length (Shannon information) requires  minimizing the Helmholtz free energy via a Helmholtz machine \citep{Hinton-wake-sleep}, which is typically computationally intractable.  
When examined closer, many commonly used mathematical or statistical ``measures'' for model goodness are either {\em exponentially expensive to compute} for general high-dimensional models or even become {\em ill-defined} for data distributions with low-dimensional supports. These measures include widely used quantities, such as maximal likelihood, KL divergence, mutual information, and Jensen-Shannon and Wasserstein distances.\footnote{More explanations about caveats associated with these measures can be found in \citep{ma2007segmentation,dai2021closedloop}.}  It is commonplace in the practice of machine learning to resort to various heuristic approximations and empirical evaluations. As a result, performance guarantees and understanding are lacking.

\begin{figure*}[t]
\centering
    \includegraphics[width=0.70\textwidth]{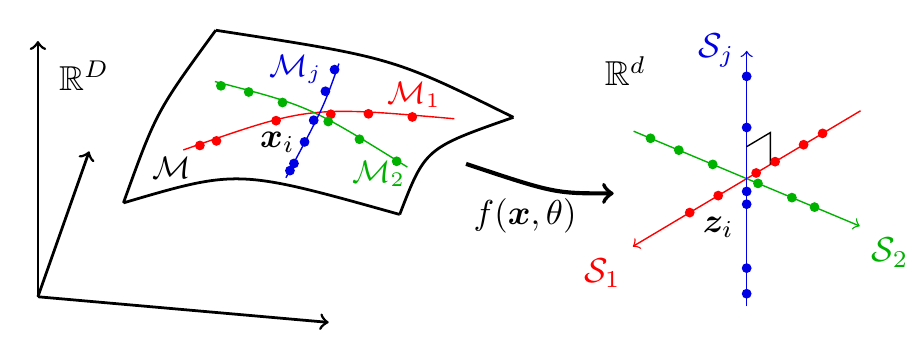}
    \caption{Seeking a linear and discriminative representation: mapping  high-dimensional sensory data, typically distributed on many  nonlinear low-dimensional submanifolds, onto a set of independent linear subspaces of the same dimensions as the submanifolds.}
     \vspace{-3mm}
    \label{fig:LDR}
\end{figure*}

Now we face a question: how can an intelligent system embody the Principle of Parsimony to identify and represent structures in observations in a computationally tractable and even efficient way? Theoretically, an intelligent system could use any family of desirable structured models for the world,
provided they are simple yet expressive enough to model informative structures in real-world sensory data. The system should be able to accurately and efficiently evaluate how good a learned model is, and the measure used should be basic, universal, and tractable to compute and optimize. What is a good choice for a family of structured models with such a measure?  

To see how we can model and compute  parsimony, we use the motivating and intuitive example of modeling visual data.\footnote{It is arguably true that vision is the most complex to model among all senses.} To make our exposition easy, we will start with a supervised setting in this section. Nevertheless, as will be discussed in the next section, with parsimony as the only  ``self-supervision'' and with the second principle of self-consistency, a learning system can become fully autonomous and function without needing any exterior supervision.

\paragraph{Modeling and computing parsimony.} Let us use $\x$ to denote the input sensory data, e.g., an image, and $\z$ its internal representation. The sensory data sample $\x \in \Re^D$ is typically rather high-dimensional (millions of pixels) but has extremely low-dimensional intrinsic structures.\footnote{For example, all images of a rotating pen trace out only a one-dimensional curve in the space of millions of pixels.} Without loss of generality, we may assume that it is distributed on some low-dimensional submanifolds, as illustrated in Figure \ref{fig:LDR}. Then, the purpose of learning is to establish a (usually nonlinear)  mapping $f$, say in some parametric family  $\theta \in \Theta$, from $\x$ to a much lower-dimensional representation $\z \in \Re^d$:
\begin{equation}
\x \in \Re^D \xrightarrow{\hspace{2mm} f(\x, \theta)\hspace{2mm}} \z \in \Re^d,
\end{equation}
such that the distribution of feature $\z$ is much more compact and structured. Being compact means economic to store. Being structured means efficient to access and use. Particularly, linear structures are ideal for interpolation or extrapolation.

\begin{figure*}
    \centering
    \includegraphics[width=0.65\textwidth]{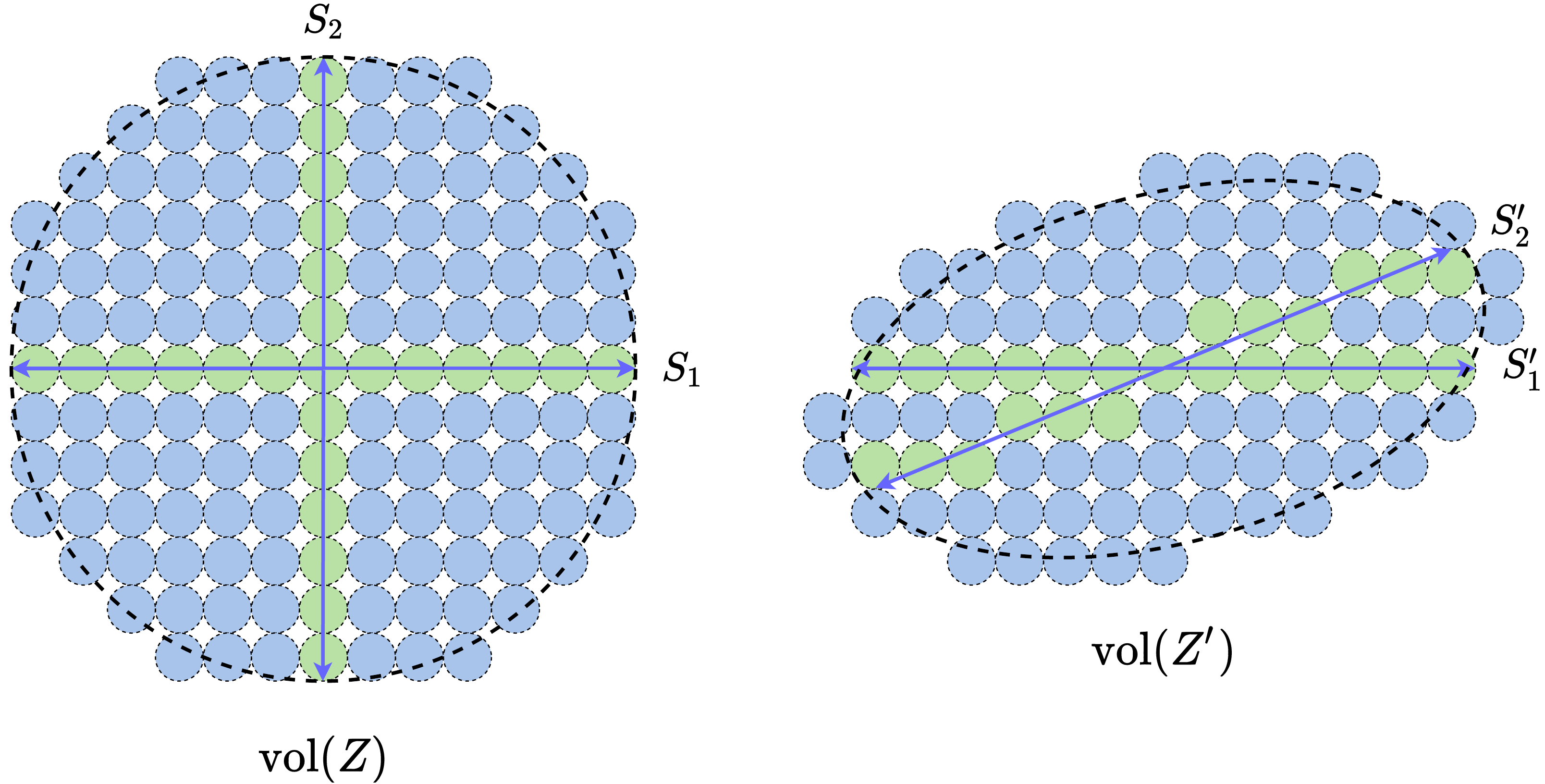}
    \caption{Rate of all features $R$ = $\log  \#(\text{\textcolor{applegreen}{green spheres} + \textcolor{cornflowerblue}{blue spheres}})$; average rate of features on the two subspaces $R^c$ = $\log  \#(\text{\textcolor{applegreen}{green spheres}})$; rate reduction is the difference between the two rates: $\Delta R = R - R^c$.}
    \vspace{-3mm}
    \label{fig:pack}
\end{figure*}

To be more precise, we can formally instantiate the principle of Parsimony for visual data modeling as trying to find a (nonlinear) transform $f$ that achieves the following goals:
\begin{itemize}
    \item \textbf{compression:} map high-dimensional sensory data $\x$ to a low-dimensional representation $\z$;
    \item \textbf{linearization:} map each class of objects distributed on a nonlinear submanifold to a linear subspace;
    \item \textbf{sparsification:} map different classes into subspaces with independent or maximally incoherent bases.\footnote{This is related to the notion of sparse dictionary learning \citep{zhai2020complete} or independent component analysis (ICA) \citep{Hyvarinen1997a,Hyvarinen1997-NC}. Once the bases of the subspaces are made independent or incoherent by the transform, the resulting representation becomes sparse and thus collectively compact and structured. For example, two sets of subspaces with the same dimensions have the same intrinsic complexity. However, their extrinsic representations can be very different, see Figure \ref{fig:pack}. This illustrates why simply compressing data based on their intrinsic complexity is insufficient for parsimony.}

\end{itemize}
In other words, we try to transform real-world data that may lie on a family of low-dimensional submanifolds in a high-dimensional space onto a family of independent low-dimensional linear subspaces. Such a model is called a {\em linear discriminative representation} (LDR) \citep{yu2020learning,chan2021redunet}, and the compression process is illustrated in Figure \ref{fig:LDR}. In some sense, one may even view the common practice of deep learning that maps each class to a ``one-hot'' vector as seeking a very special type of LDR models in which each target subspace is only one-dimensional and orthogonal to others. 

The idea of compression as a guiding principle of the brain for representing (sensory data of) the world has strong roots in neuroscience, going back to Barlow's efficient coding hypothesis \citep{Barlow1961-ce}. Scientific studies have shown that visual object representations in the brain exhibit compact structures,  such as sparse codes \citep{olshausen1996emergence} and subspaces \citep{Chang-Cell-2017,Bao2020AMO}. This supports the proposal that low-dimensional linear models are the preferred representations in the brain (at least for visual data). 

\paragraph{Maximizing rate reduction.} Remarkably, for the family of LDR models, there is a natural intrinsic measure of parsimony. Intuitively speaking, given an LDR, we can compute the total ``volume'' spanned by all features on all subspaces and the sum of ``volumes'' spanned by features of each class. Then the ratio between these two volumes gives a natural measure that suggests how good the LDR model is: the larger, the better. Figure \ref{fig:pack} shows an example with features distributed on two subspaces, $S_1$ and $S_2$. Models on the left and right have the same intrinsic complexity. The configuration on the left is preferred as features for different classes are made independent and orthogonal -- their extrinsic representations would be the most sparse. Hence, in terms of this basic volumetric measure, the best representation should be such that ``{\em the whole is maximally greater than the sum of its parts}.''

As per information theory, the volume of a distribution can be measured by its {\em rate distortion} \citep{Thomas-Cover}. Roughly speaking, the rate distortion is the logarithm of how many $\epsilon$-balls or spheres one can pack into the space spanned by a distribution.\footnote{Sphere packing gives almost a universal way to measure the volume of space of arbitrary shape: to compare volumes of two containers, one only has to fill them both with beans and then count and compare.
Optimal sphere-packing problems can be traced back to Johannes Kepler since 1611. Most recently, Mathematician Maryna Viazovska received the 2022 Fields medal for solving the optimal sphere packing problem in spaces of dimension 8 and 24, respectively \cite{sphere-packing-8,sphere-packing-24}!
}  The logarithm of the number of balls directly translates into how many binary bits one needs in order to encode a random sample drawn from the distribution subject to the precision $\epsilon$. This is generally known as the {\em description length} \citep{Rissanen,ma2007segmentation}. 

Now let $R$ be the rate distortion of the joint distribution of all features $\Z \doteq  [\z^1, \ldots, \z^n]$ of sampled data $\X \doteq [\x^1, \ldots, \x^n]$ from all, say $k$, classes. $R^c$ is the average of the rate distortions for the $k$ classes: $R^c(\Z) = \frac{1}{k} [R(\Z_1) + \cdots + R(\Z_k)]$ where $\Z = \Z_1 \cup \cdots \cup \Z_k$. Note that, because of the logarithm, the ratio between volumes becomes the difference between rates. Then the difference between the whole and the sum of the parts is called the {\em rate reduction} \citep{chan2021redunet}:
\begin{equation}
    \Delta R(\Z) \doteq R(\Z) - R^c(\Z),
    \label{eqn:rate-reduction}
\end{equation}
gives the most basic, bean-counting-like, measure of how good the feature representation $\Z$ is.\footnote{The rate reduction quantity also has a natural interpretation as ``information gain'' \citep{decision-trees}. It measures how much information is gained, in terms of bits saved, by specifying a sample on one of the parts, compared to drawing a random sample from the whole.}

Although for general distributions in high-dimensional spaces, the rate distortion, like many other measures mentioned before, is intractable and NP-hard to compute \citep{rate-distortion}, the rate distortion for data $\Z$ drawn from a Gaussian supported on a subspace has a closed-form formula \citep{ma2007segmentation}:
\begin{equation}
R(\Z) \doteq \frac{1}{2}\log\det\left(\I + \alpha\Z\Z^{*}\right).
\label{eqn:rate-distortion}
\end{equation}
Hence, it can be efficiently computed and optimized! 

\begin{figure*}
\centering
    \includegraphics[width=0.70\textwidth]{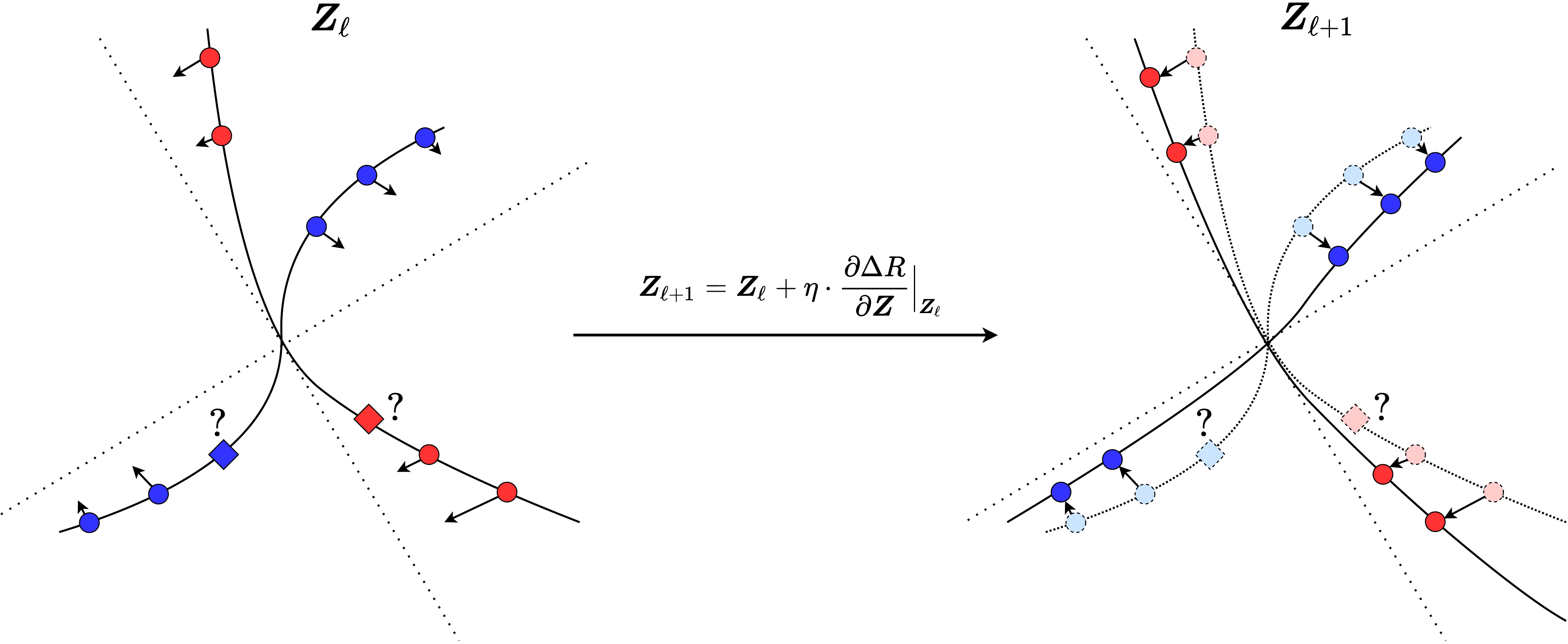} 
    \caption{A basic way to construct the nonlinear mapping $f$: following the local gradient flow $\frac{\partial \Delta R(\Z)}{\partial \Z}$ of the rate reduction $\Delta R$, we incrementally linearize and compress features on nonlinear submanifolds and separate different submanifolds to respective orthogonal subspaces (the two dotted lines).}
    \label{fig:deform}
    \vspace{-3mm}
\end{figure*} 
The work of \cite{chan2021redunet} has shown that if one uses the rate distortion functions of Gaussians and chooses a generic deep network (say a ResNet) to model the mapping $f(\x,\theta)$, then by maximizing the coding rate reduction, known as {\em the MCR$^2$ principle}:
\begin{equation}
    \max_\theta \Delta R(\bm Z(\theta)) = R(\Z(\theta)) - R^c(\Z(\theta)),
    \label{eqn:MCR2}
\end{equation}
one can effectively map a multi-class visual dataset to multiple orthogonal subspaces. Notice that maximizing the first term of the rate reduction $R$ expands the volume of all features. It simultaneously conducts ``constrastive learning'' for all features, which can be much more effective than contrasting sample pairs as normally conducted in conventional contrastive methods \citep{hadsell2006dimensionality,oord2018representation}. Minimizing the second term $R^c$ compresses and linearizes the features in each class. This can be interpreted as conducting ``contractive learning'' \citep{contractive-ICML11} for each class. The rate reduction objective unifies and generalizes these heuristics.

Particularly, one can rigorously show that, by maximizing the {rate reduction}, features of different classes will be independent and features of each class will be distributed {\em almost uniformly} within each subspace \citep{chan2021redunet}. In contrast, the widely practiced {\em cross entropy} objective  for mapping each class to a one-hot label maps the final features of each class onto a one-dimensional singleton \citep{papyan2020prevalence}.

\paragraph{White-box deep networks from unrolling optimization.}

Notice that in this context, the role of a deep network is simply to model the nonlinear mapping $f$ between the external data $\x$ and the internal representation $\z$. How should an intelligent system know what family of models to use for the map $f$ in the first place? Is there a way to directly derive and construct such a mapping instead of guessing and trying different possibilities? 

Recall that our goal is to optimize the rate reduction $\Delta R(\Z)$ as a function of the set of features $\Z$. To this end, we may  directly start with the original data $\Z_0 = \X$ and incrementally optimize $\Delta R(\Z)$, say with a {\em projected gradient ascent} (PGA) scheme:\footnote{For fair comparison of coding rates  between two representations, we need to normalize the scale of the features, say $\|\z\| =1$.}
\begin{eqnarray}
\bm Z_{\ell+1}   \propto \bm Z_{\ell} + \eta \cdot \frac{\partial \Delta R}{\partial \bm Z}\bigg|_{\Z_\ell}
 \; \mbox{subject to} \; \|\Z_{\ell+1}\| = 1.
\label{eqn:gradient-descent}
\end{eqnarray}
That is, one can follow the gradient of the rate reduction to move the features so as to increase the rate reduction. Such a gradient-based iterative deformation process is illustrated in Figure \ref{fig:deform}.

\begin{figure*}
\centering
    \includegraphics[width=5cm]{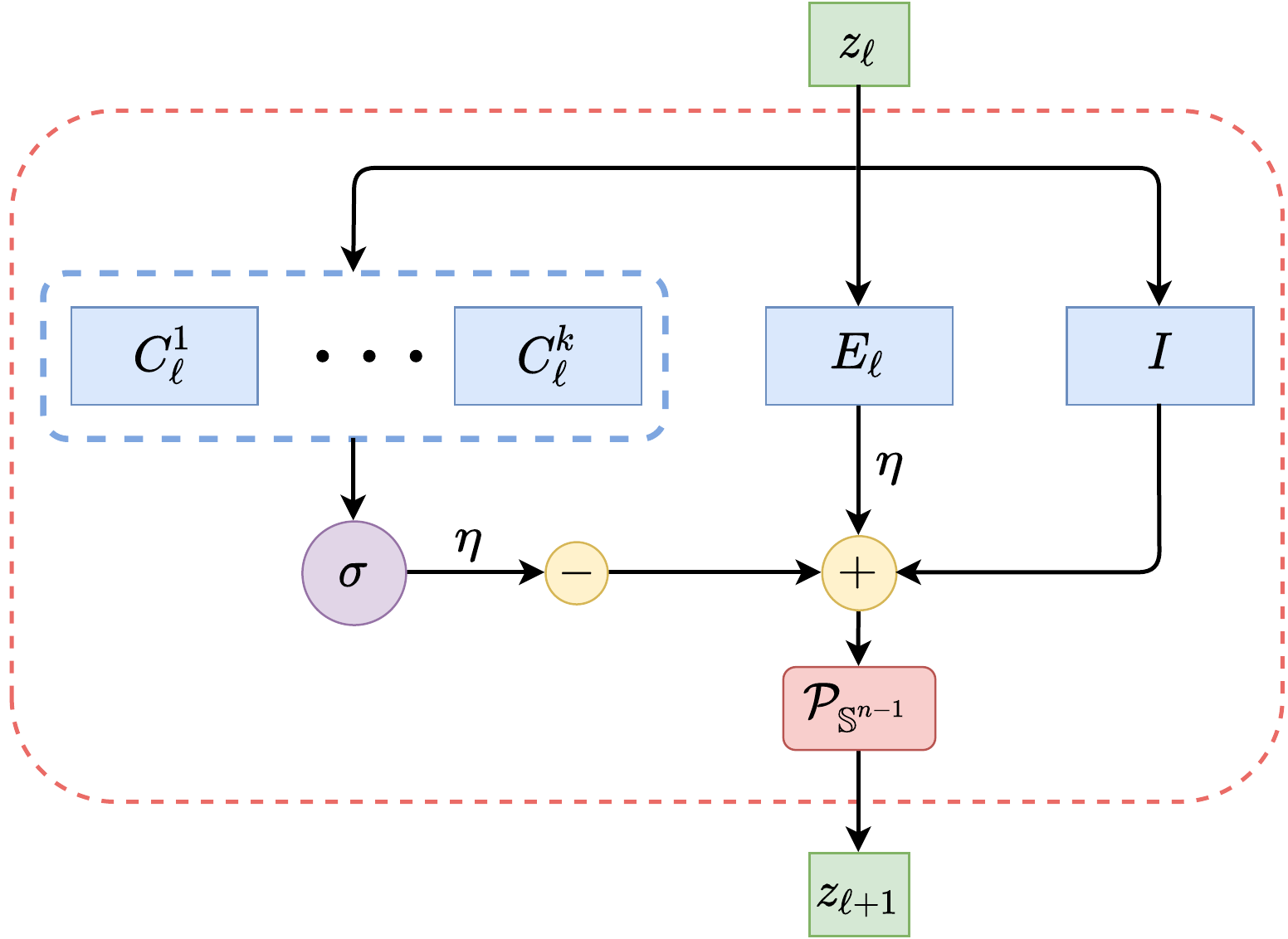} \hspace{5mm} \includegraphics[width=10.5cm]{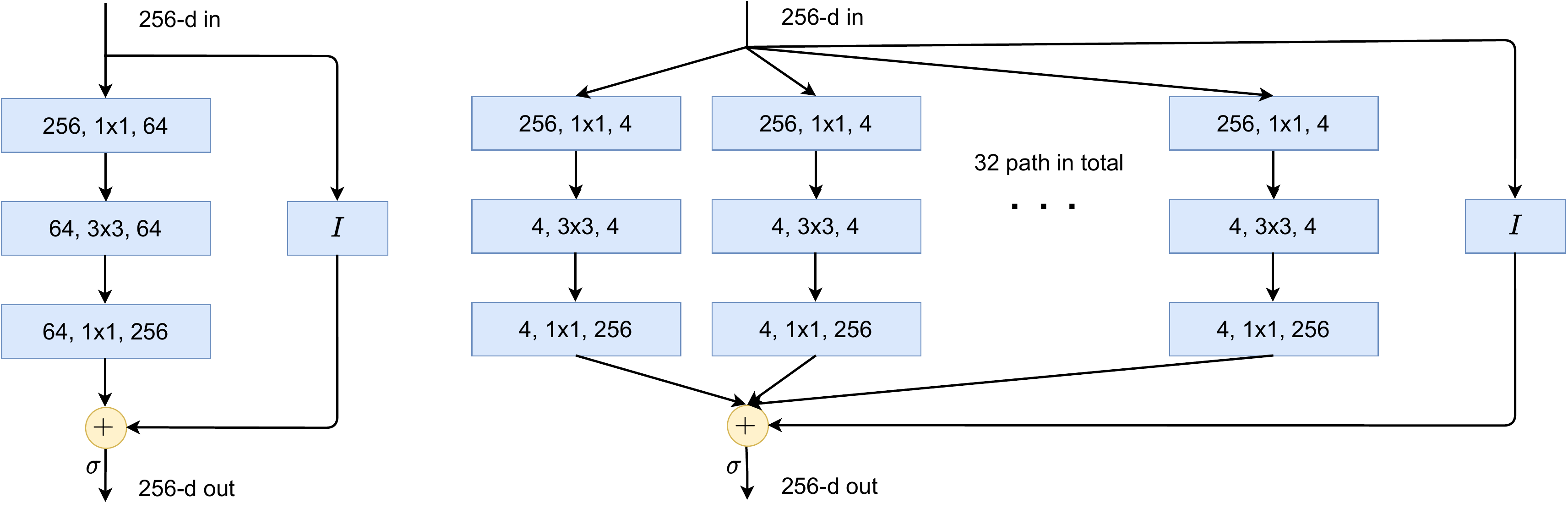}
    \caption{Building blocks of the nonlinear mapping $f$. Left: one layer of the \textbf{ReduNet} as one iteration of projected gradient ascent, which precisely consists of expansive or compressive linear operators, a nonlinear softmax, plus a skip connection, and normalization. Middle and Right: one layer of \textbf{ResNet}  and \textbf{ResNeXt}, respectively.}\vspace{-3mm}
    \label{fig:ReduNet}
\end{figure*}

From the closed-form formula for the rate distortions \eqref{eqn:rate-distortion}, we can also compute the gradient of $\Delta R = R - R^c$ in closed-form. For example, the gradient of the first term $R$ is of the form:
\begin{eqnarray}
\frac{\partial R(\Z)}{\partial \Z}\bigg|_{\Z_\ell} &=& \frac{1}{2}\frac{\partial \log \det (\I \!+\! \alpha \Z \Z^{*} )}{\partial \bm Z}\bigg|_{\Z_\ell} \\
&=& \alpha(\I \!+\! \alpha\Z_\ell \Z_\ell^{*})^{-1}\Z_\ell \doteq \E_{\ell} \Z_\ell.
\label{eqn:gradient}
\end{eqnarray}
Similarly, we can compute the gradients for the $k$ terms $\{R(\Z_i)\}_{i=1}^k$ in $R^c$ and obtain $k$ operators on $\Z_\ell$, named as $\C_i$.
Then, the above gradient ascent operation \eqref{eqn:gradient-descent} takes the following structured form:
\begin{eqnarray}
\z_{\ell+1}  &\propto&   \z_\ell +  \eta \cdot \Big[ \bm E_{\ell} \z_{\ell} +  \bm \sigma\big([\bm{C}_{\ell}^{1} \z_{\ell}, \dots, \bm{C}_{\ell}^{k} \z_{\ell}]\big)\Big] \nonumber \\ 
&&\mbox{subject to} \quad \|\z_{\ell +1}\|=1,
\label{eqn:layer-approximate}
\end{eqnarray}
where $\bm E_\ell$ and $\bm C_\ell$'s are linear operators fully determined by covariances of the features from the previous layer $\Z_{\ell}$ \eqref{eqn:gradient}.\footnote{$\bm E$ is associated with the gradient of the first term $R$ and stands for ``expansion'' of the whole set of features, whereas $\bm C$'s are associated with the gradients of multiple rate distortions in the second term $R^c$ and stand for ``compression'' of features in each class. See \cite{chan2021redunet} for the details.} Here, $\bm \sigma$ is a softmax operator that assigns $\z_\ell$ to its closest class based on its distance to each class, measured by $\bm  C_\ell \z_\ell$. A  diagram of all the operators per iteration is given in Figure \ref{fig:ReduNet} left. 

Acute readers may have recognized that such a diagram draws a good resemblance to a layer of popular ``tried-and-tested''  deep networks such as ResNet  \citep{he2016deep} (Figure \ref{fig:ReduNet} middle), including parallel columns as in ResNeXt \citep{ResNEXT} (Figure \ref{fig:ReduNet} right) and a Mixture  of Experts (MoE) \citep{MoE}. This provides a natural and plausible  interpretation of an important class of deep neural networks from the perspective of {\em unrolling an optimization scheme}. Even before the rise of modern deep networks, iterative optimization schemes for seeking sparsity, such as ISTA or FISTA \citep{Wright-Ma-2021},  had been interpreted as learnable deep networks, e.g., the work of \cite{gregor2010learning} on Learned ISTA.\footnote{A strong connection between sparsity and deep  convolution neural networks (CNNs) is formally established by \cite{sparse-land}. Similarly, unfolding iterative optimization for sequential sparse recovery leads to recurrent neural networks (RNNs)  \citep{wisdom2017building}.} The class of networks derived from optimizing rate reduction has been named as {\em ReduNet} \citep{chan2021redunet}.

\paragraph{Forward unrolling versus backward propagation.} We see above that compression leads to an entirely constructive way of deriving a deep neural network, including its architecture and parameters, as a fully interpretable {\em white-box}\footnote{Here, we only give an interpretation of deep networks, instead of AI in general, which, we believe, remains an open research topic, as we will discuss more in Section \ref{sec:broad-program}.}: its layers conduct iterative and incremental optimization of a principled objective that promotes parsimony. As a result, for so-obtained deep networks, the ReduNets, starting from the data $\X$ as input, each layer's operators and parameters $(\E_\ell, \C_\ell)$ are constructed and initialized in an entirely {\em forward unrolling} fashion. This differs from the popular practice  in deep learning: starting with a randomly constructed and initialized network which is then tuned globally via backward propagation \citep{Back-Prop}. It is widely believed that the brain is unlikely to utilize backward propagation as its learning mechanism due to the requirement for symmetric synapses and the complex form of feedback. Here, the forward unrolling optimization only relies on operations between adjacent layers that can be hard-wired; hence, it would be much easier for nature to realize and exploit. 

Additionally, parameters and operators of the so-constructed networks are amenable to further fine-tuning via another level of optimization, e.g., (stochastic) gradient descent realized by backward propagation \citep{Back-Prop}.\footnote{It has been shown that the ReduNets have the same model capacity (say to interpolate all training data precisely) as tried-and-tested deep networks such as ResNets \citep{chan2021redunet}.} But one should not confuse the (stochastic) gradient descent used to fine-tune a network with the gradient-based optimization that layers of the network ought to realize.

\paragraph{CNN derived from shift-invariance and nonlinearity.} If we further wish  the learned encoding $f$ to be {\em invariant} (or equivariant) to all time-shifts or space-translations, then we view every sample $\x(t)$ with all its shifted versions $\{\x(t-\tau)\; \forall \tau\}$ as in the same equivalence class. If we compress and linearize them together into the same subspace, then all the linear operators, $\bm E$ or $\bm C$'s, in the above gradient operation \eqref{eqn:layer-approximate} automatically become {\em multi-channel convolutions} \citep{chan2021redunet}! As a result, the ReduNet naturally becomes a multi-channel convolution neural network (CNN), originally proposed for shift-invariant recognition \citep{Fukushima,LeNet-1}.\footnote{In addition, due to special structures in such convolution operators $\E$ and $\C$'s, they are much more efficient to be computed in the {\em frequency domain} than in the time/space domain: the computational complexity reduces from $O(D^3)$\footnote{Notice that computing $\E_{\ell}$ requires inverting a $D\times D$ matrix $\alpha(\I \!+\! \alpha\Z_\ell \Z_\ell^{*})^{-1}$ which is in general of complexity $O(D^3)$.} to $O(D)$ in the dimension $D$ of the input signals \citep{chan2021redunet}.}



\paragraph{Artificial selection and evolution of neural networks.} Once we realize the role of the deep networks themselves is to conduct (gradient-based) iterative  optimization to compress, linearize, and sparsify data, it may become easy to understand the ``evolution'' of artificial neural networks that has occurred in the past decade. Particularly, it helps explain why only a few have emerged on top through a process of {\it
artificial selection}: going from general MLPs to CNNs to ResNets to Transformers. In comparison, a random search of network structures, such as Neural Architecture Search \citep{NAS-1,Baker2017DesigningNN} and AutoML \citep{automl}, has not resulted in any network architecture that is effective for general tasks.  We speculate that successful architectures are simply getting more and more effective and flexible at emulating iterative optimization schemes for data  compression. Besides the aforementioned similarity between ReduNet and ResNet/ResNeXt, we want to discuss a few more examples.

\paragraph{What is a transformer transforming?} Notice that the gradient of a rate distortion term $R(\Z)$ is of the form \eqref{eqn:gradient}: $\frac{\partial R}{\partial \Z} = \alpha (\I + \alpha\Z_\ell \Z_\ell^{*})^{-1}\Z_\ell$. Instead of viewing the matrix $\alpha (\I + \alpha\Z_\ell \Z_\ell^{*})^{-1}$ as a linear operator $\E_{\ell}$ acting on $\Z_\ell$, as was  done in the ReduNet, we may rewrite the whole gradient term approximately as:
\begin{eqnarray}
\alpha (\I + \alpha\Z_\ell \Z_\ell^{*})^{-1}\Z_\ell\! &\!\approx\!&\! \alpha (\I  - \alpha\Z_\ell \Z_\ell^{*})\Z_\ell \nonumber \\
\!&\!=\!&\! \alpha \big[\Z_\ell - \alpha \Z_\ell (\Z_\ell^*\Z_\ell)\big].
\label{eqn:transformer-gradient}
\end{eqnarray}
That is, the gradient operation for optimizing a rate distortion term depends mainly on the auto-correlation of the features $\bm A \doteq \Z_\ell^*\Z_\ell$ from the previous iteration. This is also known as ``self-attention'' or ``self-expression'' in some contexts \citep{attention,vidal2022attention}. If we consider applying an additional {\em learnable}  linear transform $\U$ to each feature term in the above expression \eqref{eqn:transformer-gradient} for the gradient, a gradient-based iteration to optimize rate distortion takes the general form:
\begin{equation}
\Z_{\ell +1}  \doteq  \Z_\ell + \U_o \big[\Z_\ell - \alpha \U_v\Z_\ell (\U_k\Z_\ell)^*(\U_q\Z_\ell)\big].
\label{eqn:transformer}
\end{equation}
This is of { exactly the same} form as the basic operation of each layer for a Transformer \citep{attention}, i.e., a self-attention (SA) head followed by a feed-forward residual MLP operation.\footnote{If the term in the bracket $\big[\Z_\ell - \alpha \U_v\Z_\ell (\U_k\Z_\ell)^*(\U_q\Z_\ell)\big]$ is interpreted as to emulate the gradient \eqref{eqn:transformer-gradient} of rate distortion, the linear operator $\bm U_o$ can then be viewed to emulate a certain regularized gradient-based method. For instance, it can be used to model the inverse of the Fisher information matrix in the natural gradient descent \citep{kakade2001natural}. } 

Moreover, very similar to ResNeXT versus ResNet, for tasks such as image classification, it is found empirically better to use {\em multiple}, say $k$, SA heads in parallel in each layer \citep{dosovitskiy2021an}. In the context of rate reduction, these SA heads may be naturally interpreted as gradient terms associated with the multiple rate distortion terms in the rate reduction $\Delta R(\Z) = R(\Z) - \big[R(\Z_1) + \cdots + R(\Z_k)\big]/k$. The learned linear transforms $(\U_k, \U_q, \U_v)$ in each SA head can be interpreted as ``matched filters'' or ``sparsifying dictionaries''\footnote{Interested readers may see  \citep{zhai2020complete} for more details about the topic of sparse dictionary learning.} that select and transform token sets (on submanifolds) that belong to the same category (of signals or images). Hence, we conjecture that layers of Transformer \eqref{eqn:transformer} emulate a more general family of gradient-based iterative schemes that optimize the rate reduction of all input token sets (on multiple submanifolds) by clustering, compressing, and linearizing them altogether.

Furthermore, gradient ascent or descent is the most basic type of optimization scheme. Networks based on unrolling such schemes (e.g., ReduNet)  might not be the most efficient yet. One could anticipate that more advanced optimization schemes, such as accelerated gradient descent methods \citep{Wright-Ma-2021}, could lead to more efficient deep network architectures in the future. Architecture wise, these accelerated methods require the introduction of {\em skip connections} across multiple layers. This may help explain, from an optimization perspective, why additional skip connections have often been found to improve network efficiency in practice, e.g., in highway networks \citep{srivastava2015highway} or dense networks \citep{huang2017densely}.



\subsection{How to Learn: the Principle of Self-Consistency}
\label{sec:self-consistency}

\begin{quote}
``{\em Everything should be made as simple as possible, but not any simpler}.''\vspace{-2mm}

$~$\hfill -- Albert Einstein
\vspace{-2mm}
\end{quote}

The principle of Parsimony alone does not ensure that a learned model will capture all important information in the data sensed about the external world. For example, mapping each class to a one-dimensional  ``one-hot'' vector, by minimizing the cross entropy, may be viewed as a form of being parsimonious. It may learn a good classifier, but the features learned would collapse to a singleton, known as {\em neural collapse} \citep{papyan2020prevalence}. The so  learned features would no longer contain enough information to regenerate the original data. Even if we consider the more general class of LDR models, the rate reduction objective  alone does not automatically determine  the correct dimension of the ambient feature space. If the feature space dimension is too low, the model learned will under-fit the data; if it is too high, the model might over-fit.\footnote{The first expansive or contrastive term in the rate reduction might over-expand the features to fill the space, due to noises or other variations.} 

More generally, we take the view that perception is distinct from the performance of specific tasks, and the goal of perception is to learn {\it everything} predictable about what is sensed. In other words, the intelligent system should be able to {\it regenerate the  distribution of the observed data from the compressed representation} to the point that itself cannot distinguish internally despite its best effort. 
This view  distinguishes our framework from existing ones that are customized to a specific class of tasks. Representative of such is the {\em information bottleneck} framework  \citep{Tishby-ITW2015,ShwartzZiv2017OpeningTB,michael2018on}, which explains how only information in the data related to its class label is extracted via the deep networks. 
To govern the process of learning a fully faithful representation\footnote{Although in this section, for simplicity, we focused our discussions on modeling 2D imagery data, we will discuss the perception of the 3D world in Section \ref{sec:vision-graphics}, as well as argue why perception needs to integrate recognition,  reconstruction, and regeneration.}, we introduce a second principle:
\vspace{-4mm}
\paragraph{The Principle of Self-consistency:} {\em An autonomous intelligent system seeks a most  self-consistent model for observations of the external world by minimizing the internal discrepancy between the observed and the regenerated.}
\vspace{2mm}

The principles of Self-consistency and Parsimony are highly complementary and should always be used together. The principle of Self-consistency alone does not ensure any gain in compression or efficiency. Mathematically and computationally, it is easy and even trivial to fit any training data with over-parameterized models\footnote{Having a photographic memory is not intelligence. It is the same as fitting all the data in the world with a Big Model.} or to ensure consistency by establishing one-to-one mappings between domains with the same dimensions without learning intrinsic structures in the data distribution.\footnote{That is the case with many popular methods for learning generative models of data, such as normalizing flows \citep{Kobyzev_2021}, cycle GAN \citep{zhu2017unpaired}, and diffusion probabilistic models \citep{diffusion}, etc. Although so learned models might be useful for applications such as image generation or style transfer, they do not identify low-dimensional structures in the data distributions nor produce compact linear structures in the learned representations.} Only through compression can an intelligent system be compelled to discover intrinsic low-dimensional structures within the high-dimensional sensory data, and transform and represent them in the feature space in the most compact way for future use. Also, only through compression can we easily understand why over-parameterization, e.g., by feature lifting with hundreds of channels, as normally done in DNNs, will not lead to over-fitting if its sheer purpose is to compress in the higher-dimensional feature space: lifting helps reduce the nonlinearity in the data\footnote{Say, as in the scattering transforms \citep{scattering-net} or random filters \citep{chan2015pcanet,chan2021redunet}.}, rendering it easier to compress and linearize.\footnote{As Lao Tzu famously said in Tao Te Ching:  ``{\em That which shrinks must first expand}.'' } The role of subsequent layers is to perform compression (and linearization), and in general, the more layers, the better it is compressed.\footnote{This naturally  explains a seemingly mystery about deep networks: the ``double-descent'' phenomenon suggests a deep model's test error becomes smaller as it gets larger, after reaching its peak at certain interpolation point \citep{belkin2019reconciling,yang2020rethinking}.}

So far, we have established that a mechanism is needed to determine if the compressed representation contains {\it all} the information that is sensed. In the remainder of this section, we will first introduce a general architecture for achieving this, a {\it generative model}, which can regenerate a sample from its compressed representation. Then, a difficult problem arises: how to sensibly measure the discrepancy between the sensed sample and the regenerated sample? We argue that for an autonomous system, there is one and only one solution to this: measuring their discrepancies in the internal feature space. Finally, we argue that the compressive encoder and the generator must learn together through a zero-sum game. Through these deductions, we derive a universal framework for learning that we believe is inevitable.

\paragraph{Auto-encoding and its caveats with computability.} To ensure the learned feature mapping $f$ and representation $\z$ have correctly captured low-dimensional structures in the data, one can check if the compressed feature $\z$ can reproduce the original data $\x$, by some generating map $g$, parameterized by $\eta$:
\begin{equation}
\x\in \Re^D \xrightarrow{\hspace{2mm} f(\x, \theta)\hspace{2mm}} \z \in \Re^d \xrightarrow{\hspace{2mm} g(\z, \eta)\hspace{2mm}} \hat{\x} \in \Re^D,
\label{eqn:autoencoding}
\end{equation}
in the sense that $\hat \x = g(\z, \eta)$ is close to $\x$ (according to a certain measure). This process is generally known as {\em auto-encoding} \citep{Kramer1991NonlinearPC,auto-encoding}. In the special case of compressing to a {\em structured} representation such as LDR, we call such an auto-encoding a {\em transcription}\footnote{This is analogous to the memory-forming transcription process of Engram \citep{Josselyn2020MemoryER} or that between functional proteins and DNA (genes).} \citep{dai2021closedloop}.  However, this goal is  easier said than done. The main difficulty lies in how to make this goal computationally tractable and hence physically realizable. More precisely, what is a principled measure for the difference  between the distribution of $\x$ and that of $\hat \x$ that is both  {\em mathematically well-defined} and {\em efficiently computable}? As  mentioned before, when dealing with distributions in high-dimensional spaces with degenerate low-dimensional supports, which is almost always the case with real-world data \citep{ma2007segmentation,Vidal:Springer16}, conventional measures, including the KL divergence, mutual information, Jensen-Shannon distance, Helmholtz free energy, and Wasserstein distances, can be either ill-defined or intractable to compute, even for Gaussians (with support on subspaces) and their mixtures\footnote{Many existing methods formulate their objectives based on these quantities. Thus, these methods typically rely on expensive brute-force sampling to approximate these quantities or optimize their approximated lower-bounds or surrogates, such as in variational  auto-encoding (VAE) \citep{kingma2013auto}. The fundamental limitations of these methods are often disguised by good empirical results obtained using clever heuristics and excessive computational resources.}. How can we resolve this fundamental and yet often unacknowledged difficulty in computability associated with comparing degenerate distributions in high-dimensional spaces?

\begin{figure*}
    \centering
    \includegraphics[width=0.9\textwidth]{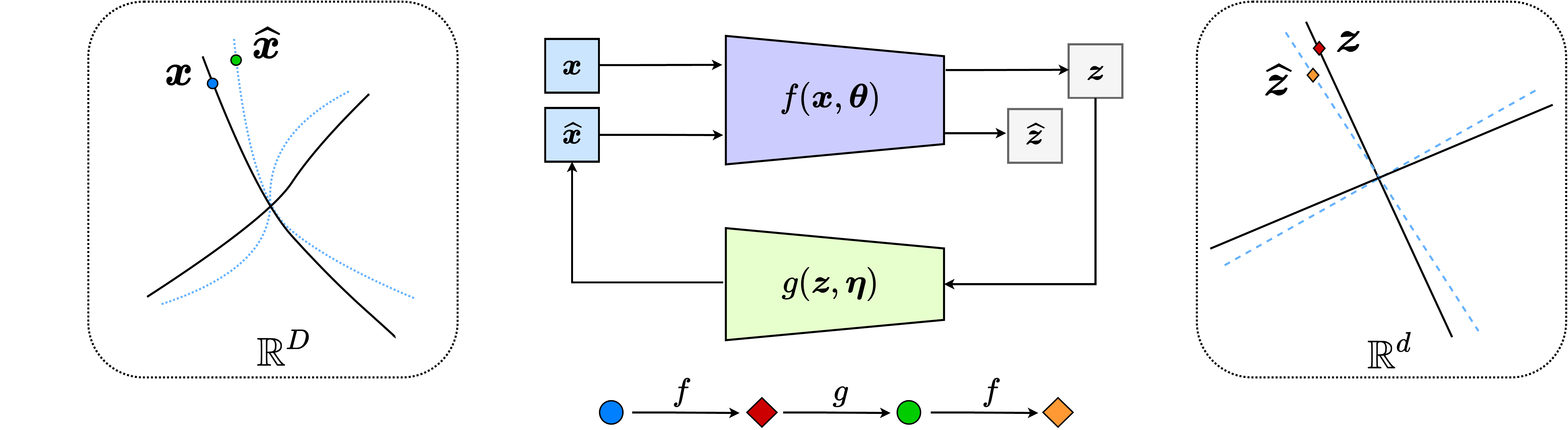}
    \caption{A compressive closed-loop transcription of nonlinear data submanifolds to an LDR, by comparing and minimizing the difference in $\z$ and $\hat \z$, internally. This leads to a natural pursuit-evasion game between the encoder/sensor $f$ and the decoder/controller $g$, allowing the distribution of the decoded $\hat{\x}$ (the dotted blue curves) to chase and match that of the observed data $\x$ (the solid black curves).} 
    \vspace{-3mm}
    \label{fig:closedloop}
\end{figure*}
\paragraph{Closed-loop data transcription for  self-consistency.} As shown in the previous section, the rate reduction $\Delta R$ gives a well-defined principled distance  measure between degenerate   distributions. However, it is computable (with closed-form) only for a  mixture of subspaces or Gaussians, not for general distributions! Yet, we can only expect the distribution of the internally structured representation $\z$ to be a  mixture of subspaces or Gaussians, not the original data $\x$.

This leads to a rather profound question regarding learning a ``self-consistent'' representation: to verify the correctness of an internal model for the external world, {\em does an autonomous agent really need to measure any discrepancy in the data space?} The answer is actually no. The key is to realize that, to compare $\x$ and $\hat \x = g(\z, \eta)$, the agent only needs to compare their respective internal features  $\z = f(\x,\theta)$ and $\hat \z = f(\hat \x,\theta)$ via the same mapping $f$ that intends to make $\z$ compact and structured. 
\begin{equation}
\x \xrightarrow{\hspace{2mm} f(\x, \theta)\hspace{2mm}} \z \xrightarrow{\hspace{2mm} g(\z, \eta)\hspace{2mm}} \hat{\x} \xrightarrow{\hspace{2mm} f(\x, \theta)\hspace{2mm}} \hat{\z}.    
\label{eqn:closed-loop}
\end{equation}
Measuring distribution differences in $\z$ space is  well-defined and efficient: it is arguably true that in the case of natural intelligence, learning to measure discrepancies {\em internally} is the only thing the brain of a self-contained autonomous agent can  do.\footnote{Imagining someone  colorblind, it is unlikely his/her internal representation of the world requires minimizing discrepancies in RGB values of the visual inputs $\x$. } 

This effectively leads to a ``closed-loop'' feedback system, and the overall process is illustrated in Figure \ref{fig:closedloop}.  The encoder $f$ now plays an additional role as a discriminator, detecting any discrepancy between $\x$ and $\hat \x$ through the difference between their internal features $\z$ and $\hat \z$. The distance between the distribution of $\z$ and that of $\hat \z$ can be measured through the rate reduction \eqref{eqn:rate-reduction} of their samples $\Z(\theta)$ and $\hat \Z(\theta, \eta)$:
\begin{equation*}
 \Delta R\big(\Z(\theta),\hat{\Z}(\theta,\eta)\big) \doteq R\big(\Z \cup \hat{\Z}\big) - \frac{1}{2}\big(R(\Z) + R(\hat{\Z})\big).
\end{equation*}

One can interpret popular practices for learning either a DNN classifier $f$ or a generator $g$ alone as learning an open-ended segment of the closed-loop system (Figure \ref{fig:closedloop}). 
This currently popular practice is very similar to an open-loop control which has long been known in the control community to be problematic and costly. The training of such an open segment requires supervision on the desired output (e.g., class labels), and deployment of such an open-loop system is inherently not stable, robust, or adaptive if the data distributions, system parameters, or tasks change. For example, deep classification networks trained in supervised settings often suffer {\em catastrophic forgetting} if retrained for new tasks with new classes of data \citep{catastrophic}. In contrast, closed-loop systems are inherently more stable and adaptive \citep{Wiener-1948}. It has been suggested by \cite{Hinton-wake-sleep} that the discriminative and generative segments need to be combined as the ``wake'' and the ``sleep'' phases, respectively, of a complete learning process.

\paragraph{Self-learning through a self-critiquing game.}
However, just closing the loop is not enough. It is tempting to think that now we only need to optimize the generator $g$ to minimize the difference between $\z$ and $\hat\z$,\footnote{This is very similar in spirit to the ``sleep'' phase of the wake-sleep scheme proposed by \cite{Hinton-wake-sleep}: it essentially tries to ensure that the encoding (recognition) network $f$ produces a response $\hat{\z}$ to the regenerated $\hat{\x} = g(\z)$ {\em consistent} with its origin $\z$.} e.g., in terms of the rate reduction measure:
\begin{equation}
    \min_\eta \Delta R\big(\Z(\theta),\hat{\Z}(\theta,\eta)\big).
    \label{eqn:error}
\end{equation}
Note that $\Delta R(\Z, \hat{\Z}) = 0$ if $\hat{\Z} = g(f(\Z)) = \Z$. That is, the optimal set of features $\Z$ should be a ``fixed point'' of the encoding-decoding loop\footnote{This can be viewed as a generalization to the ``deep equilibrium models'' \citep{NEURIPS2019_01386bd6} or the ``implicit deep learning'' models \citep{implicit}. Both interpret deep learning as conducting fixed point computation from a feedback control perspective.}. But the encoder $f$ performs significant dimension reduction and compression, so $\hat{\Z} = \Z$ does not necessarily imply $\hat{\X} = \X$. To see this, consider the simplest case when $\X$ is already on a linear subspace (e.g., of dimension $k$) and $f$ and $g$ are a linear projection and lifting, respectively \citep{Druv-Pai}. $f$ would not be able to detect any difference in its (large) null space: $\X$ and any $\hat{\X} = \X + \mbox{null}(f)$ have the same image under $f$. 

How can $\hat{\Z} = \Z$ imply $\hat{\X} = \X$ then? In other words, how can satisfaction with the self-consistency criterion in the internal space guarantee that we have learned to regenerate the observed data faithfully? This is possible {\em only when the dimension $k$ is low enough and $f$ can be further adjusted}. Let us assume that the dimension of $\X$ is $k < d/2$, where $d$ is the dimension of the feature space. Then the dimension $\hat{\X} = g(f(\X))$ under a linear lifting $g$ is a subspace of $k$-dimension. The union of the two subspaces of $\X$ and $\hat \X$ is of dimension at most $2k < d$. Hence, if there is a difference between these two subspaces and $f$ can be an arbitrary projection, we have $f(\X) \not= f(\hat{\X})$, i.e., $\X \not= \hat{\X}$ implies $\Z \not= \hat{\Z}$.


Hence, after $g$ minimizes the error $\Delta R$ in \eqref{eqn:error}, $f$ needs to actively adjust and detect, in its full capacity, if there is remaining discrepancy between $\X$ and $\hat{\X}$, e.g., by maximizing the same measure $\Delta R$. The process can be repeated between the encoder $f$ and the decoder $g$,  resulting in a natural {\em pursuit and evasion game}, as illustrated in Figure \ref{fig:closedloop}. 

In the 1961 edition of his book {\em Cybernetics}, \cite{Wiener-1961} added a supplementary chapter discussing learning through playing games. The games he described were mostly about an intelligent agent against an opponent or the world (which we will discuss in the next section). Here we advocate the need of an {\em internal} game-like mechanism for any intelligent agent to be able to conduct self-learning via self-critique! What abides is the notion of (non-cooperative) games as a universally effective way of learning \citep{Game-von-Neumann,Game-Nash}: applying the current model or strategy repeatedly against an adversarial critique, hence continuously improving the model or strategy based on feedback received through a closed loop!

Within such a framework, the encoder $f$ assumes a dual role. In addition to learning a representation $\z$ for the data $\x$ by maximizing the rate reduction $\Delta R(\Z)$ (as done in Section \ref{sec:parsimony}), it should serve as a feedback ``sensor'' that actively detects any discrepancy between the data $\x$ and the generated $\hat \x$. Also, the decoder $g$ assumes a dual role: it is a ``controller'' that corrects any discrepancy between $\x$ and $\hat \x$ detected by $f$, as well as a decoder trying to minimize the overall coding rate $\Delta R(\hat{\Z})$ needed to achieve this goal (subject to a given precision). 

\begin{figure*}[t]
\centering
\includegraphics[width=0.85\textwidth]{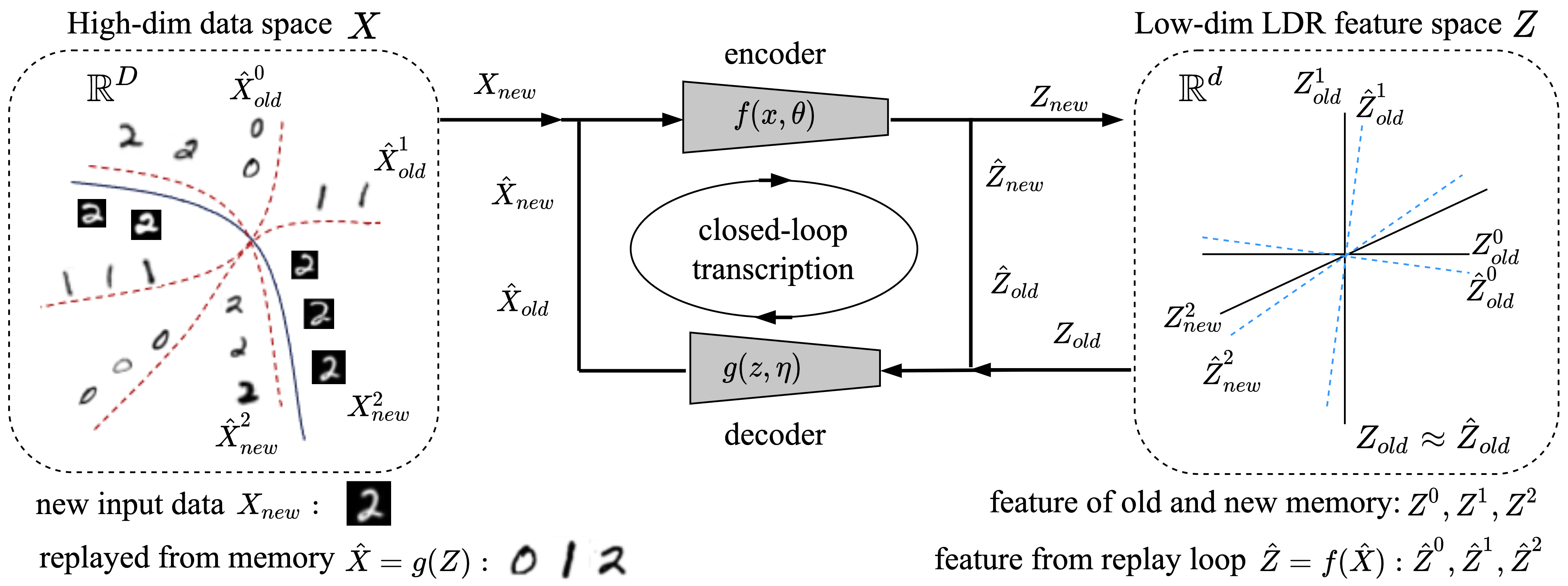}
\caption{{Incremental learning} via a compressive closed-loop transcription. For a new data class $\X_{new}$, a new LDR memory $\Z_{new}$ is  learned via a constrained minimax game between the encoder and decoder subject to a constraint that  memory of past classes $\Z_{old}$ is preserved, as a ``fixed point'' of the closed loop.}
\label{fig:framework-incremental}
\vspace{-3mm}
\end{figure*}

Therefore, the optimal ``parsimonious'' and  ``self-consistent'' representation tuple $(\z, f, g)$ can be interpreted as the {\em  equilibrium point} of a zero-sum  game between $f(\cdot, \theta)$ and $g(\cdot, \eta)$, over a combined rate reduction based utility on $\Z(\theta)$ and $\hat{\Z}(\theta, \eta)$:
\begin{equation}
   \max_\theta\min_\eta \Delta R \big(\Z \big) + \Delta R \big(\hat{\Z} \big) +  \Delta R\big(\Z,\hat{\Z}\big). 
   \label{eqn:maximin}
\end{equation}
A recent analysis has rigorously shown that, in the case when the input data $\X$ lie on multiple linear subspaces, the desired optimal representation for $\Z$ is indeed  
the Stackelberg equilibria  \citep{fiez2019convergence,jin2019local} of a sequential maximin game over a rate reduction objective similar to the above \citep{Druv-Pai}. It remains an open problem for the case when $\X$ are on multiple nonlinear submanifolds. Nevertheless, compelling empirical evidence indicates that solving this game indeed provides excellent auto-encoding for real-world visual  datasets  \citep{dai2021closedloop}, and automatically determines a subspace with a proper dimension for each class. It does not seem to suffer from problems like {\em mode collapsing} in training conventional generative models,   such as GAN \citep{veegan}. The so-learned representation is simultaneously {\em  discriminative and generative.} 




\paragraph{Self-consistent incremental and unsupervised learning.} 

So far, we have mainly discussed the two principles in the supervised setting. In fact, a primary advantage of our framework is that it is most natural and effective for {\em self-learning} via { self-supervision} and self-critique. Additionally, since the rate reduction has sought  explicit (subspace-type) representations for the learned structures,\footnote{instead of a ``hidden'' or ``latent'' representation learned using a purely generative method such as GAN \citep{goodfellow2014generative} where the features are distributed randomly in the feature space.} this makes it easy for past knowledge to be preserved  when learning new tasks/data, as a prior (memory) to be kept {\em self-consistent}.

For more clarity, let us examine how the closed-loop transcription framework above can be naturally extended to the case of {\em incremental learning} -- that is, to learn to recognize one class of objects at a time instead of simultaneously learning many classes. While learning the representation $\Z_{new}$ for a new class, one only needs to add the cost to the objective \eqref{eqn:maximin} and ensure the representation $\Z_{old}$ learned before for old classes remains {\em self-consistent} (a fixed point) through the closed-loop transcription: $\Z_{old} \approx \hat{\Z}_{old} = f(g(\Z_{old}))$. In other words, the above maximin game \eqref{eqn:maximin} becomes a game with constraints:
\begin{eqnarray}
  &\!\!\!\!\!\max_\theta\min_\eta & \!\!\!\!\!\Delta R(\Z) + \Delta R\big(\hat{\Z}\big)  + \Delta R(\Z_{new},\hat{\Z}_{new}) \nonumber \\
   &\!\!\!\!\!\!\!\!\mbox{subject to} & \!\!\!\!\!\Delta R\big(\Z_{old},\hat{\Z}_{old}\big) = 0. 
   \label{eqn:maximin-constrained}
\end{eqnarray}
Such a constrained game makes learning an incremental and dynamic process, enabling the learned transcription to  {\em adapt} to new incoming data continuously. This process is illustrated in Figure \ref{fig:framework-incremental}.

Recent empirical studies \citep{tong2022incremental} have shown that this leads to arguably the first self-contained neural system with a fixed capacity that can incrementally learn good LDR representations {\em without suffering catastrophic forgetting} \citep{catastrophic}. Forgetting, if any, is rather graceful with such a closed-loop system. Additionally, when images of an old class are provided again to the system for review, the learned representation can be further {\em consolidated} -- a characteristic very similar to that of human memory. In some sense, such a constrained closed-loop  formulation ensures that the visual memory formation can be {\em Bayesian} and {\em adaptive} -- characteristics hypothesized to be desirable for the brain \citep{FRISTON2009293}. 

\begin{figure*}[t]
     \centering
\includegraphics[width=0.265\textwidth]{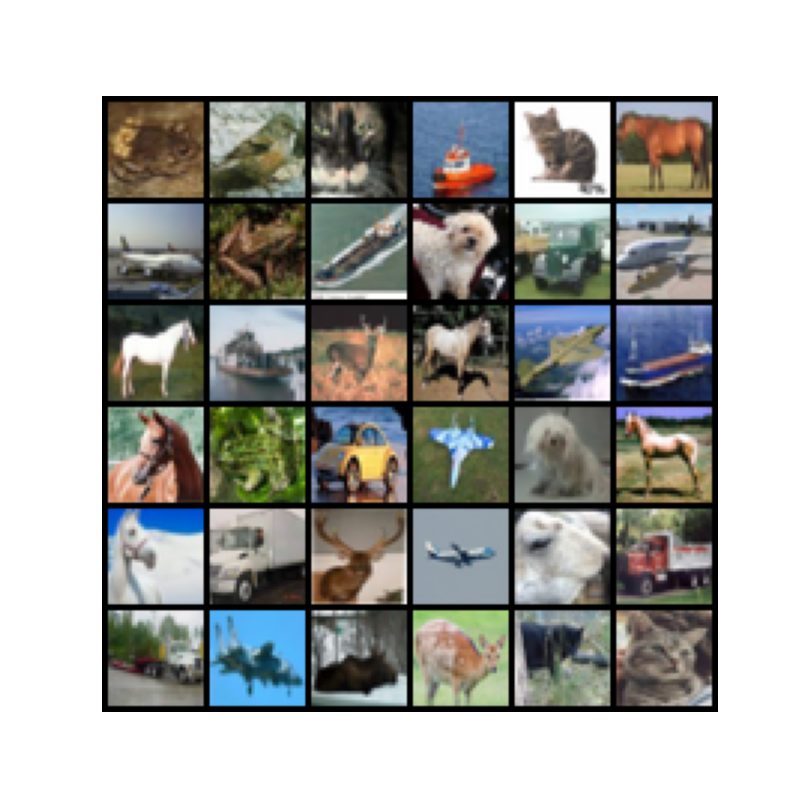} \hspace{-6mm}
\includegraphics[width=0.265\textwidth]{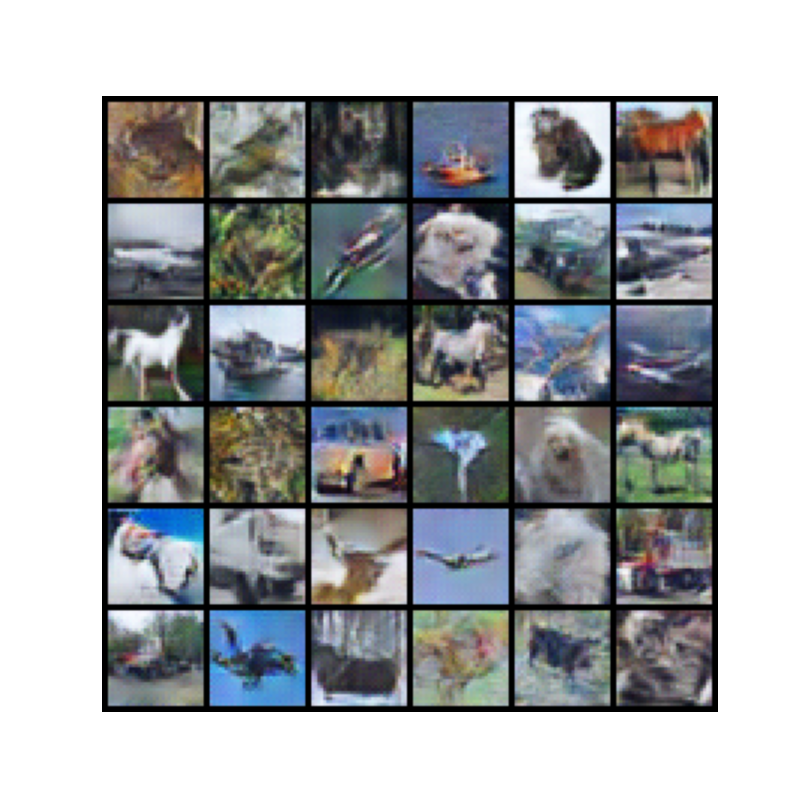} 
\includegraphics[width=0.46\textwidth]{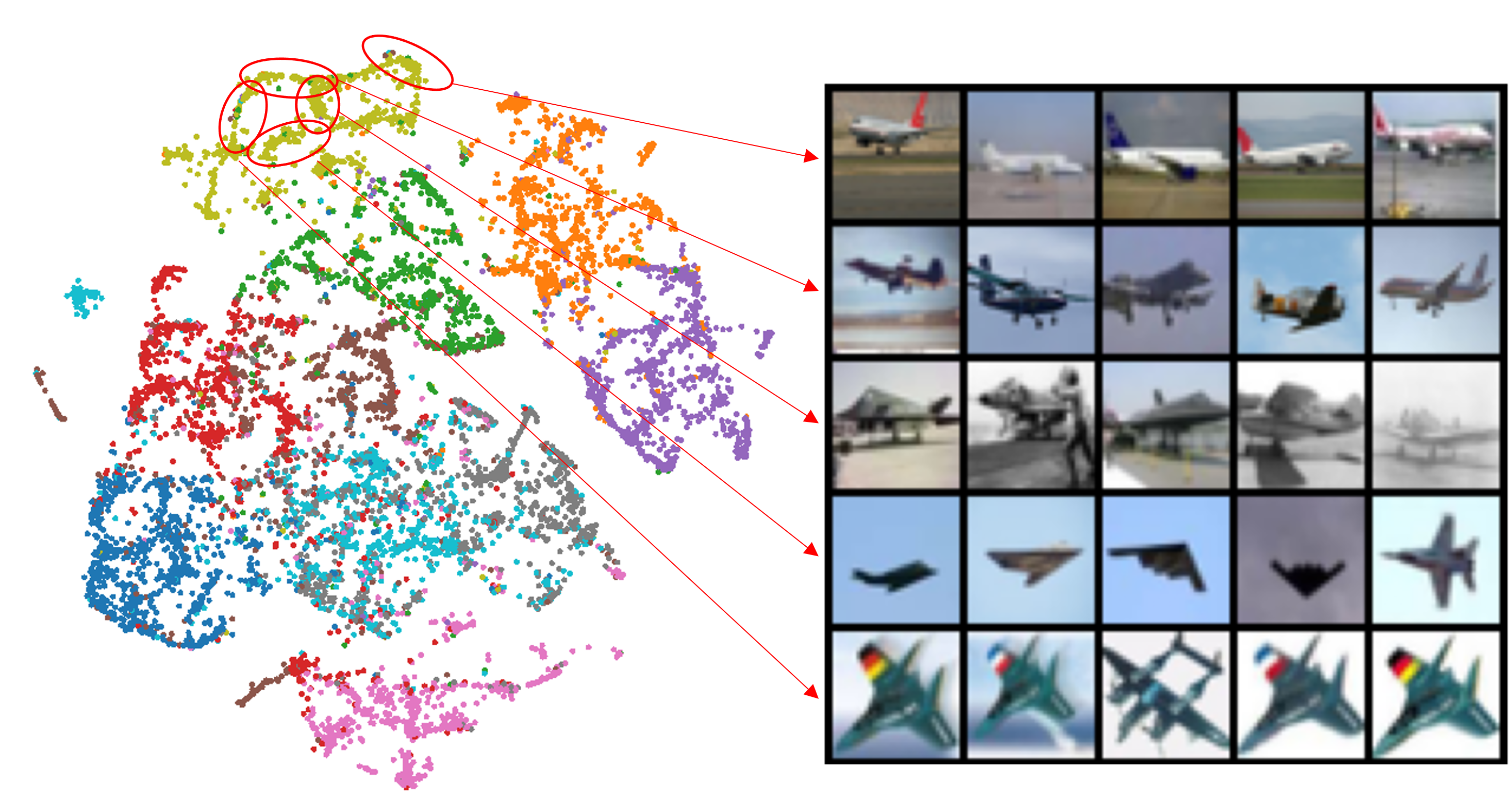}
    \caption{Left: comparison between $\x$ and the corresponding decoded $\hat \x$ of the  auto-encoding learned in the unsupervised setting for the CIFAR-10 dataset (with 50,000 images in ten classes). Right: t-SNE of unsupervised-learned features of the ten  classes and visualization of several neighborhoods with their associated images. Notice the local thin (nearly 1-D) structures in the visualized features, projected from a feature space of hundreds of dimensions.}
    \label{fig:vis_recon}
   \vspace{-2mm}
\end{figure*}

\begin{figure}
    \centering
    \includegraphics[height=5cm]{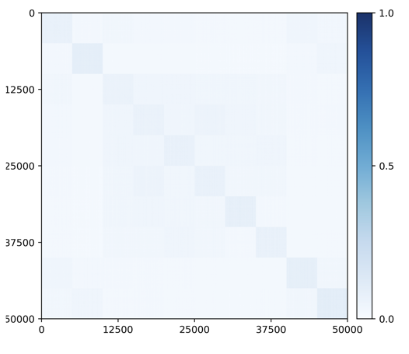}
    \caption{Correlations between unsupervised-learned features for 50,000 images that belong to ten classes (CIFAR-10) by the  closed-loop transcription. Block-diagonal structures consistent with the classes emerge without any supervision. }
    \label{fig:feature-correlation}
    \vspace{-1mm}
\end{figure}

Note that this framework is fundamentally conceived to work in an entirely unsupervised setting. Thus, even though for pedagogical purposes we presented the principles assuming that class information is available, the framework can be naturally extended to an entirely {\em unsupervised setting} in which no class information is given for any data sample. Here, we only have to view every new sample and its augmentations as one new class in \eqref{eqn:maximin-constrained}. This can be viewed as one type of ``self-supervision.'' With the ``self-critiquing'' game mechanism, a compressive closed-loop transcription can be easily learned.  As shown in Figure \ref{fig:vis_recon},  the so-learned auto-encoding shows good sample-wise consistency, and the learned features also demonstrate clear and meaningful local low-dimensional (thin) structures. More surprisingly, subspaces or block-diagonal  structures in the feature correlation emerge in the features learned for the classes even without any class information provided during training at all (Figure \ref{fig:feature-correlation})! Hence, structures of the so-learned features resemble those of category-selective areas observed in a primate's brain \citep{Kanwisher4302,Kanwisher-2010, Kriegeskorte2008-gf, Bao2020AMO}.

\section{Universal Learning Engines}\label{sec:learning-engine}
\begin{quotation}
``{\em What I cannot create, I do not understand.}''\vspace{-1mm}

$~$\hfill -- Richard Feynman
\vspace{-2mm}
\end{quotation}

In the above section, we deduced from the first principles of Parsimony and Self-consistency the compressive closed-loop transcription  framework, using the example of modeling visual imagery data. In the remaining two sections, we offer more speculative thoughts on the universality of this framework, extending it to 3D vision and reinforcement learning (the rest of this section)\footnote{Our discussions on the two topics require familiarity with certain domain specific terminology and knowledge. Readers who are less familiar with these topics may skip without much loss of continuity.}  and projecting its implications for neuroscience, mathematics, and higher-level intelligence (Section \ref{sec:broad-program}).

\paragraph{``Unite and build'' versus ``divide and conquer.''} Within the compressive closed-loop transcription framework, we have seen  why and how fundamental ideas and concepts from coding/information theory, feedback control, deep networks, optimization, and game theory all come together to become integral parts of a complete intelligent system that can learn. Although ``divide and conquer'' has long been a cherished tenet in scientific research, regarding understanding a complex system such as intelligence, the opposite ``unite and build'' should be the tenet of choice. Otherwise, we would forever be {\em blind men with an elephant}: each person would always believe a small piece  is the whole world and tend to blow its significance out of proportion.\footnote{Hence all the superficial claims: ``this or that is all you need.''} 

The two principles serve as the glue needed to combine many necessary pieces together for the jigsaw puzzle of intelligence, with the role of deep networks naturally and clearly revealed as models for the nonlinear mappings between external observations and internal representations. Interestingly, the principles reveal computational mechanisms for learning systems that resemble some of the key characteristics observed in or hypothesized about the brain, such as  sparse coding and subspace coding \citep{Barlow1961-ce,olshausen1996emergence,Chang-Cell-2017}, 
closed-loop feedback \citep{Wiener-1948}, and free energy minimization \citep{FRISTON2009293}, as we will discuss more in the next section.

Notice that closed-loop compressive architectures are ubiquitous for all intelligent beings and at all scales, from the brain (which compresses sensory information) to spinal circuits (which compress muscle movements) down to DNA (which compresses functional information about proteins). We believe compressive closed-loop transcription may be the {\em universal learning engine} behind all intelligent behavior. It enables intelligent beings and systems to discover and distill low-dimensional structures from seemingly complex and unorganized input and transform them into compact and organized internal structures for memorizing and exploitation.

To illustrate the universality of such a framework, for the remainder of this section, we examine two more tasks:  {\em 3D  perception} and {\em decision making}, which are believed to be two key modules for any autonomous intelligent system \citep{LeCun-2022}. We speculate on how, guided by the two principles, one can develop different perspectives and new insights to understand these challenging tasks.

\begin{figure*}[t]
\centering
\includegraphics[width=0.9\textwidth]{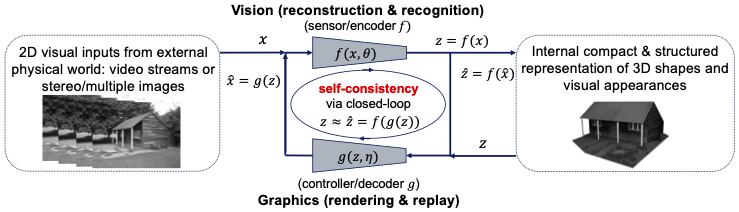}
\caption{A closed-loop relationship between Computer Vision and Graphics for a compact and structured 3D model of the visual inputs.}
\label{fig:vision-graphics}
\vspace{-3mm}
\end{figure*}

\subsection{3D Perception: Closing the Loop for Vision and Graphics} \label{sec:vision-graphics}

Thus far, we have demonstrated the success of closed-loop transcription in discovering compact structures in datasets of 2D images. This relies on the existence of {\em statistical correlations} among imagery data in each class. We believe that the same compression mechanisms would be even more effective if the low-dimensional structures in the data were defined through hard physical or geometric constraints rather than soft statistical correlations. 

Particularly, if we believe that the principles of Parsimony and Self-consistency also play a role in how the human brain develops mental models of the world from life-long visual inputs, then our sense of 3D space should be the result of such a closed-loop compression or transcription. The classic paradigm for 3D vision laid out by David Marr in his influential book {\em Vision} \citep{Marr} advocates a ``divide and conquer'' approach that partitions the task of 3D perception into several modular processes: from low-level 2D processing (e.g.,  edge detection and contour sketching), to mid-level 2.5D parsing (e.g., grouping, segmentation, and figure and ground), and high-level 3D reconstruction (e.g., pose and shape) and recognition (e.g., objects). In contrast, the compressive  closed-loop transcription proposed in this paper advocates an opposite ``unite and build'' approach.

\paragraph{Perception as a compressive closed-loop transcription?} More precisely, a three-dimensional representation of shapes, appearances, and even dynamics of objects in the world should be the most compact and structured representation our brain has developed internally to consistently interpret all perceived visual observations. If so, the two principles then suggest that a compact and structured 3D representation is directly the internal model to be sought for. This implies that we could and should unify Computer Vision and Computer Graphics within a single closed-loop computational framework, as illustrated in Figure \ref{fig:vision-graphics}.

Computer Vision has conventionally been interpreted as a forward process that reconstructs and recognizes an internal 3D model for the 2D visual inputs \citep{Ma:2003:IVI,Szeliski}, whereas Computer Graphics \citep{Computer-Graphics} represents its inverse process that renders and animates the internal 3D model. There might be tremendous computational and practical benefits to directly combining these two processes into a closed-loop system: all the rich structures (e.g.,  sparsity and smoothness) in geometric  shapes, visual appearances, and dynamics can be exploited together for a unified 3D model that is the most compact and consistent with all visual inputs. 

Indeed, the recognition techniques in computer vision could help computer graphics in building compact models in the spaces of shapes and appearance and enabling new ways for creating realistic 3D content. Conversely, the 3D modeling and simulation techniques in computer graphics could predict, learn and verify the properties and behavior of the real objects and scenes analyzed by computer vision algorithms. In fact, the approach of "analysis by synthesis" has been long practiced by the vision and graphics community, e.g., for efficient online perception \citep{yildirim2020efficient}. Some recent examples of closing the loop for computer vision and graphics include a learned 3D rendering engine \citep{NIPS2015_ced556cd} and 3D aware image synthesis \citep{chanmonteiro2020pi-GAN,Wood_2021_ICCV}. 

\paragraph{Unified representations  for appearance and shape?} Image-based rendering \citep{levoy1996light, gortler1996lumigraph,shum2011image}, in which a new view is generated by learning from a set of given images, may be regarded as an early attempt to close the gap between vision and graphics with the principles of parsimony and self-consistency. 
Particularly, plenoptic sampling \citep{chai2000plenoptic} showed that an anti-aliased image (self-consistency) can be achieved with the minimum number of images required (parsimony). 

\begin{figure*}[t]
\centering
\includegraphics[width=0.77\textwidth]{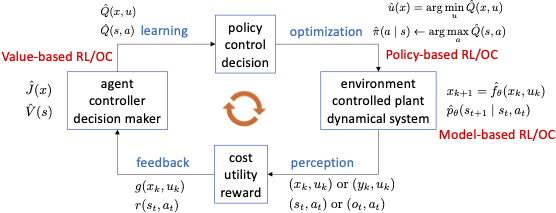}
\caption{An autonomous intelligent agent that integrates perception (feedback), learning, optimization, and action in a closed loop to learn an optimal policy for a certain task. $s_t$ or $x_k$ is the state of the world model; $r$ or $g$ is the perceived reward or cost of action $a_t$ or control $u_k$ on the current state; $J$ or $V$ is the (learned) cost or value associated with each state, $Q$ is the (learned) cost associated with each state-action pair. Here, we deliberately use terminologies from optimal control (OC)  \citep{Bertsekas-2012} and reinforcement learning (RL) \citep{Sutton-Barto} in parallel for both comparison and unification.  }
\label{fig:RL}
\vspace{-3mm}
\end{figure*}
Recent developments in modeling radiance fields have provided more empirical evidence for this view \citep{yu2021plenoxels}: directly exploiting low-dimensional structures in the radiance field in 3D (sparse support and spatial smoothness) leads to much more efficient and effective solutions than brute-force training of black-box deep neural networks \citep{mildenhall2020nerf}. However, it remains a challenge for the future to identify the right family of compact and structured 3D  representations that can integrate  shape geometry, appearance, and even dynamics in a unified framework that leads to minimal complexity in data, model, and computation.

\subsection{Decision Making: Closing the Loop for Perception, Learning, and Action}
Thus far in this paper, we have discussed how compressive closed-loop transcription may lead to an effective and efficient framework for learning a good perceptual model from visual inputs. At the next level, an autonomous agent can use such a perceptual model to achieve certain tasks in a complex {\em dynamical} environment. The overall process for the agent to learn from perceived results or received rewards for its actions forms another closed loop at a higher level (Figure \ref{fig:RL}). 

The principle of self-consistency is clearly at play here: the role of the closed-loop feedback system is to ensure that the learned model and control policy by the agent is {\em consistent} with the external world in such a way that the model can make the best prediction of the state ($s_t$) transition, and the learned control policy $\pi_\theta$ for the action ($a_t$) results in maximal expected reward $R$\footnote{In many practices of RL, people may  consider a ``narrower'' version of the self-consistency principle: it only requires the learned state model and control policy consistent with a specific task or reward, not a full state model for all sensed data.}:
\begin{equation}
   \max_{\theta}  R(\theta) \doteq  \mathbb E_{a_t\sim \pi_\theta(s_t)} \Big[\sum_t r(s_t, a_t) \Big].
   \label{eqn:RL}
\end{equation}
Note here that the reward $R$ plays a similar role as the rate reduction objective \eqref{eqn:MCR2} for LDR models, which measures the ``goodness'' of the learned control policy $\pi$ and guides its improvement.

The principle of Parsimony is the main reason for the  success of modern reinforcement learning in tackling large-scale tasks such as Alpha-Go \citep{silver2016mastering,silver2017mastering} and playing video games \citep{berner2019dota,vinyals2019grandmaster}. In almost all tasks that have a state-action space of astronomical size or dimension, e.g., $D$, practitioners always assume that the optimal value function $V^*$, Q-function $Q^*$, or policy $\pi^*$ only depends on a small number of, e.g., $d \ll D$,  features:
\begin{eqnarray}
    V^*(\bm s) &\approx&  \hat{V}\big(f(\bm s, \bm a)\big), \nonumber\\ 
    Q^*(\bm s, \bm a) &\approx&  \hat{Q}\big(f(\bm s, \bm a)\big), \\
    \pi^*(\bm a\mid \bm s) &\approx& \hat{\pi}\big(\bm a; f(\bm s, \bm a)\big), \nonumber
\end{eqnarray}
where $f(\bm s,\bm a) \in \Re^d$ is a nonlinear mapping that learns some low-dimensional features of the extremely large or high-dimensional state-action space. In the case of video games, the state dimension $D$ is easily in the millions  and yet the number of features $d$ needed to learn a good policy is typically only a few dozen or hundred! Very often, these optimal control policies or value/reward functions sought in OC/RL are even assumed to be a linear superposition of these features \citep{ICML2000-Ng,NIPS2001-Kakade}:
\begin{equation}
    \bm \omega^\top f(\bm s, \bm a) = \omega_1\cdot f_1(\bm s, \bm a) + \cdots + \omega_d\cdot f_d(\bm s, \bm a). 
    \label{eqn:linear-regression}
\end{equation}
That is, the nonlinear mapping $f$ is assumed to be able  also to {\em linearize} the dependency of the policy/value/reward functions on the learned features.\footnote{It is a common practice in systems theory to linearize any nonlinear dynamics before controlling them,  through either nonlinear mappings known as the Koopman operators \citep{Koopman1931HamiltonianSA} or feedback linearization \citep{Sastry}.}


\paragraph{Autonomous feature selection via a game?} Notice that all these practices in RL are very similar in spirit to the learning objectives under the principle of Parsimony stated in Section \ref{sec:parsimony}. Effectively exploiting the low-dimensional structures is the (only) reason the learning can be so {\em scalable} with such a high-dimensional state-action space, and correctly identifying and linearizing such low-dimensional structures is the key for the so-learned control policy to be {\em generalizable}\footnote{Otherwise, the learned model/policy tends to over-fit or under-fit.}. Nevertheless, a proper choice in the number of features $d$ remains heuristically designed by the human in practice. That makes the overall RL not autonomous. We believe that, for a closed-loop learning system to automatically determine the right number of features associated with a reward/task, one must extend the RL  formulation \eqref{eqn:RL} to a certain maximin game\footnote{by introducing a self-critique of the features selected and learned.}, in a similar spirit as those studied in Section \ref{sec:self-consistency} for achieving  Self-consistency for visual modeling.

\paragraph{Data and computational efficiency of RL?} Recently, there have been many theoretical attempts to explain the empirically observed efficiency of reinforcement learning in terms of sampling and computation complexity of Markov decision processes (MDP). However, any theory based on unstructured generic MDPs and reward functions would not be able to provide pertinent explanations to such empirical successes. For example, some of the best known bounds on the sample complexity for reinforcement learning remain linear in cardinality of the state space and action $O(|\mathcal S||\mathcal A|)$ \citep{NEURIPS2020_Chen}, which does not explain the empirically observed efficiency of RL in large-scale tasks (such as Alpha-Go  and video games) where the state or action spaces are astronomical.

We believe that the efficiency of RL in tackling many  practical large-scale tasks  can come {\em only} from the intrinsic low-dimensionality in the system dynamics or correlation between the optimal policy/control and  the states. For example, assuming the systems have a bounded eluder dimension \citep{Eluder-dim} or the MDPs are of low rank \citep{uehara2021representation,agarwal2020flambe}. Deep networks' role is again to identify and model such low-dimensional structures and hopefully linearize them. 

To conclude, for large-scale RL tasks, the two principles together make such a closed-loop system of perception, learning, and action a truly efficient and effective learning engine. With such an engine, autonomous agents can discover low-dimensional structures if there are indeed such structures in the environment and the learning task, and eventually act intelligently when the structures learned are good enough and generalize well!


\section{A Broader Program for Intelligence}\label{sec:broad-program}
\begin{quotation}
\vspace{-1mm}
``{\em If I were to choose a patron saint for cybernetics out of the history of science, I should have to choose Leibniz.  The philosophy of Leibniz centers about two closely related concepts -- that of a universal symbolism and that of a calculus of reasoning.}''

$~$\hfill -- Norbert Wiener, {\em Cybernetics}, 1961
\end{quotation}

It has been ten years since the dramatic revival of deep neural networks with the work of \cite{krizhevsky2012imagenet}, which has garnered  considerable enthusiasm for artificial intelligence in both the technology industry and the scientific community. Subsequent theoretical studies of deep learning often view deep networks themselves as the object of study \citep{PDLT-2022}. However, we argue here  that deep networks are better understood  as a means to an end: they clearly arise to serve the purposes of identifying and transforming nonlinear low-dimensional structures in high-dimensional data, a universal task for learning from high-dimensional data  \citep{Wright-Ma-2021}.

More broadly, in this paper, we have proposed and argued that Parsimony and Self-consistency are two fundamental principles responsible for the emergence of intelligence, artificial or natural. The two principles together lead to a closed-loop computational framework that unifies and clarifies many practices and empirical findings of deep learning and artificial intelligence. Furthermore, we believe that they will 
guide us from now on to study intelligence with a more principled and integrated approach. Only in doing so can we achieve a new level of understanding the science and mathematics behind intelligence. 


\subsection{Neuroscience of Intelligence}  One would naturally expect any fundamental principle of intelligence to have major implications for the design of the most intelligent thing in the universe, the brain.  As already mentioned, the principles of Parsimony and Self-consistency shed new light on several experimental observations concerning the primate visual system. Even more importantly, they shine light on what to look for in future experiments.

We have shown that seeking an internally parsimonious and predictive representation alone is enough ``self-supervision'' to allow structures to emerge automatically in the final representation learned through a compressive closed-loop transcription. For example, Figure \ref{fig:feature-correlation} shows that unsupervised data transcription learns features that automatically distinguish different categories, providing an explanation for category-selective representations observed in the brain \citep{Kanwisher4302,Kanwisher-2010, Kriegeskorte2008-gf, Bao2020AMO}.  
These features also provide a plausible explanation for the widespread observations of sparse coding \citep{olshausen1996emergence} and subspace coding \citep{Chang-Cell-2017, Bao2020AMO} in the primate's  brain. Furthermore, beyond the modeling of visual data, recent studies in neuroscience suggest the emergence of other structured representations in the brain, such as ``place cells'', might also be the result of coding spatial information in the most compressed way \citep{Benna2021}.

Arguably, the maximal coding rate reduction (MCR$^2$) principle \eqref{eqn:MCR2} is similar in spirit to the ``free energy minimization  principle'' from cognitive science \citep{FRISTON2009293}, which attempts to provide a framework for Bayesian inference through energy minimization. But unlike the general notion of free energy, the rate reduction is computationally tractable and directly optimizable as it can be expressed in closed form. Furthermore, the interplay of our  two principles suggests that autonomous learning of the correct  model (class) should be conducted via a closed-loop maximin game over such a utility \eqref{eqn:maximin}, instead of minimization alone. Thus, we believe that our framework provides a new perspective on how to practically implement Bayesian inference.


Our framework clarifies the overall learning architecture used by the brain. One important insight is that a feed-forward segment can be constructed by unrolling an optimization scheme; learning from a random network via back propagation is unnecessary. Furthermore, our framework suggests the existence of a complementary generative segment to form a closed-loop feedback system to guide learning. Finally, our framework sheds light on the elusive ``prediction error'' signal sought by many neuroscientists interested in brain mechanisms for ``predictive coding,'' a computational scheme with resonances to compressive closed-loop transcription \citep{Rao1999-pj, Keller2018-ez}. For reasons of computational tractability, the discrepancy between incoming and generated observations should be measured at the final stage of representation.

So far, many resemblances between this new framework and the natural intelligence discussed in this paper are still speculative and remain to be corroborated with future scientific experiments and evidence. Nevertheless, these speculations offer neuroscientists ample new perspectives and hypotheses about why and how intelligence could emerge in nature. 

\subsection{Mathematics of Intelligence}
In terms of mathematical or statistical models for data analysis, one can view our framework as a generalization of PCA \citep{Jolliffe1986}, GPCA \citep{Vidal:Springer16}, RPCA \citep{candes2011robust}, and Nonlinear PCA \citep{Kramer1991NonlinearPC} to the case with multiple low-dimensional nonlinear submanifolds in a high-dimensional space. These classical methods largely model data with single or multiple linear subspaces or with a single nonlinear submanifold. We have argued that the role of deep networks is mainly to model the nonlinear mappings that simultaneously linearize and separate multiple low-dimensional submanifolds. This generalization is necessary to connect these idealistic,  classic models to the true structures of the real-world data. Despite promising and exciting empirical evidence, the mathematical properties of the compressive closed-loop transcription process remain understudied and poorly understood.  To the best of our knowledge, only for the case when the original data $\x$ are assumed to be on multiple linear subspaces, it has been shown that the maximin game based on rate reduction yields the correct optimal solution \citep{Druv-Pai}. Little is known about the transcription of nonlinear submanifolds.

A rigorous and systematic investigation requires an understanding of high-dimensional probability distributions with low-dimensional supports on submanifolds \citep{Manifold-hypothesis}. Hence,  mathematically, it is crucial to study how such submanifolds in high-dimensional spaces can be identified, grouped or separated, deformed and flattened with minimal distortion to their original metric, geometry, and topology \citep{isomap,multiple-manifolds,wang2021deep,Shamir2021TheDM}. Problems like these seem to fall into an understudied area between classical Differential Geometry and Differential Topology in mathematics. 

Additionally, we often also wish that during the process of deformation, the probability measure of data on each submanifold is redistributed in a certain optimal way such that coding and sampling will be the most economical  and efficient. This is related to topics  such as Optimal Transport \citep{Optimal-Transport}. For the case when the submanifolds are fixed linear subspaces, understanding the achievable extremals of the rate reduction, or ratio of volumes of the whole and the parts, seems related to certain fundamental inequalities in analysis such as the {\em Brascamp-Lieb} inequalities \citep{Brascamp-Lieb-1}. More general problems like these seem to be related to the studies of Metric Entropy (also commonly known as sphere packing) and Coding Theory for distributions over more general compact structures or spaces. 

Besides nonlinear low-dimensional structures, real-world data and signals are typically {\em invariant} to shift in time, translation in space, or to more general group transformations. \cite{Wiener-1961} recognized that simultaneously dealing with both nonlinearity and invariance  presents a major technical challenge. He had made early attempts to generalize Harmonic Analysis to nonlinear processes and systems.\footnote{He used his analysis to explain brain waves \citep{Wiener-1961}!} Indeed, the revival of deep learning has reignited strong interest in this critical problem, and significant  progress has been made recently, including the work of \cite{scattering-net,sparse-land,Wiatowski-2018,CohenW16,cohen2019general}. Our framework suggests that a more unifying approach to dealing with nonlinearity and invariance is through (incremental) compression. That has led to a natural derivation of structured deep networks such as the convolution ReduNet \citep{chan2021redunet}. We believe that compression provides a unifying perspective to modeling general {\em sequential} data or processes with nonlinear dynamics too, which could lead to mathematically rigorous justification for popular models such as RNNs or LSTMs \citep{LSTM}.

But besides pure mathematical interests, we must require  that the mathematical investigation leads to {\em computationally tractable} measures and {\em scalable} algorithms. One must characterize the precise statistical and computational resources needed to achieve such tasks, in the same spirit as the research program set for Compressive Sensing  \citep{Wright-Ma-2021}, because intelligence needs to apply them to model  high-dimensional data and solve large-scale tasks. This entails to ``close the loop'' between Mathematics and Computation, enabling the use of rich families of good  geometric structures (e.g., sparse codes, subspaces, grids, groups, or graphs; Fig. \ref{fig:framework}, right) as compact archetypes for modeling real-world data, through efficiently computable nonlinear mappings that generalize deep networks, e.g.,   \cite{Bronstein2021-un}. 

\subsection{Toward Higher-Level Intelligence} The two principles laid out in this paper are mainly for explaining the emergence of intelligence in  {\em individual} agents or systems, related to the notion of {\em ontogenetic learning} that Nobert Wiener first proposed \citep{Wiener-1948}. It is probably noncoincidental that after more than seventy years, we find ourselves in this paper ``closing the loop'' of the  modern practice of Artificial Intelligence back to its roots in Cybernetics and interweaving the very same set of fundamental concepts that Wiener touched upon in his book while conducting inquiries into the jigsaw puzzle of intelligence: {\em  compact coding of information, closed-loop feedback, learning via games, white-box modeling, nonlinearity, shift-invariance, etc.} 

As shown in this paper, the compressive closed-loop transcription is arguably the first computational framework that coherently integrates many of these pieces  together. It is in close  spirit to earlier frameworks \citep{Hinton-wake-sleep} but makes them computationally tractable and scalable. Particularly, the learned nonlinear encoding/decoding mappings of the loop, often  manifested as deep networks, essentially provide a much needed ``interface'' between external unorganized raw sensory data (say visual, auditory etc.) and internal compact and structured representations. 

However, the two principles proposed in this paper do not necessarily explain all aspects of intelligence. Computational mechanisms behind the emergence and development of high-level semantic, symbolic, or logic inferences remain elusive, although many foundational works have been set forth by pioneers like John McCarthy, Marvin Minsky, Allen Newell and Herbert Simon since the 1950s \citep{Simon-Book,Newell-Simon} and a comprehensive modern exposition can be found in \cite{russel2020}. To date, there remain active and contentious debates about whether such high-level symbolic intelligence can emerge from continuous learning or must be hard-coded \citep{DBLP:journals/corr/abs-2002-06177, LeCun-Browning}.

In our view, structured internal representations, such as subspaces, are necessary intermediate steps for the emergence of high-level semantic or symbolic concepts--each subspace corresponds to one {\em discrete} category (of objects). For example, notion of eyes and months may come out naturally from observing a large number of face images.  Additional statistical, causal or logical  relationships among the so-abstracted discrete concepts can be further modeled parsimoniously as a compact and structured (say sparse) graph, with each node representing a subspace/category, e.g., \citep{bear2020learning}. We believe such a graph can be and should be learned via a closed-loop transcription to ensure self-consistency. 

We conjecture that only on top of {\em compact and structured} representations learned by individual agents can the emergence and development of high-level intelligence (with shareable symbolic knowledge) be possible, subsequently and eventually. 
We suggest that new principles for the  emergence of high-level intelligence, if any, should be sought through the need for efficient communication of information or transfer of knowledge among intelligent agents. This is related to the notion of {\em phylogenetic learning} that Wiener also discussed \citep{Wiener-1961}. Furthermore, any new principle needed for such higher-level intelligence must reveal {\em reasons} why alignment and sharing of internal concepts across different individual agents is computationally possible, as well as reveal certain measurable {\em gains} in intelligence for a group of agents from such symbolic abstraction and sharing. 

\paragraph{Intelligence as interpretable and computable systems.}
Obviously, as we advance our inquiries into higher-level intelligence, we want to set much higher standards this time. Whatever new principles that might remain to be  discovered in the future, for them to truly play a substantial role in the emergence and development of intelligence, they must share two characteristics with the two principles we have presented in this paper: 
\vspace{-1mm}
\begin{enumerate}
    \item \textbf{Interpretability:} {\em all principles together should help reveal computational mechanisms of intelligence as a white box\footnote{Again, the phrase ``white box'' modeling has been conveniently borrowed from Wiener's {\em Cybernetics} \citep{Wiener-1961}.}, including measurable objectives, associated computing architectures, and structures of learned representations.}
    \vspace{-1mm}
    \item \textbf{Computability:} {\em any new principle for intelligence must be computationally tractable and scalable, physically realizable by computing machines or nature, and ultimately corroborated with scientific evidence.}
\end{enumerate}
Only with such fully interpretable and truly realizable principles in place can we explain all existing intelligent systems, artificial or natural, as partial or holistic instantiations of these principles. Then, they can help us discover effective architectures and systems for different intelligent tasks without relying on the current expensive and time-consuming  ``trial-and-error'' approach to advance. Also, we will be able to characterize the minimal data and computation resources needed to achieve these tasks, instead of the current brute-force approach that advocates ``the bigger, the better.'' Intelligence should not be the privilege of the most resourceful, as it is not the way of nature. Instead, parsimony and autonomy are the main  characteristics.\footnote{A tiny ant is arguably much more intelligent and independent than any legged robot in the world, with merely a quarter of a million neurons consuming negligible energy.} Under a correct set of principles, anyone should be able to design and build future generations of intelligent systems, small or big, with autonomy, ability, and adaptiveness, eventually emulating and even surpassing that of animals and humans. 

\section{Conclusion}
Through this paper, we hope to have convinced the reader that we are now at a much better place than people like Wiener and Shannon seventy years ago regarding  uncovering, understanding, and exploiting the works of intelligence. We have proposed and argued that, under the two principles of Parsimony and Self-consistency, it is possible to assemble many necessary pieces of the puzzle of intelligence into a unified computational framework that is easily implementable on machines or by nature. This unifying framework offers new perspectives on how we could  further advance the study of perception, decision making, and intelligence in general. 

To conclude to our proposal for a principled approach to intelligence, we emphasize once again that all scientific principles for intelligence should not be philosophical guidelines or conceptual frameworks formulated or developed with mathematical quantities that are intractable to compute or can only be approximated heuristically. They should rely on the most basic and principled objectives that are measurable with finite observations and lead to computational systems that can be realized even with limited resources. This belief is probably best expressed through a quote from Lord Kelvin:\footnote{that we have learned from Professor Jitendra Malik of UC Berkeley.} 
\begin{quotation}
``{\em When you can measure what you are speaking about and express it in numbers, you know something about it; but when you cannot measure it, when you cannot express it in numbers, your knowledge is of the meager and unsatisfactory kind: it may be the beginning of knowledge, but you have scarcely, in your thoughts, advanced to the stage of science, whatever the matter may be.}'' 

$~$\hfill -- Lord Kelvin,  1883
\end{quotation}

\section*{Afterword and Acknowledgment} Although the research of Yi and Harry focuses more on computer vision and computer graphics, they both happened to major in control and automation as undergraduate students. They started their collaboration many years ago at Microsoft Research Asia (MSRA) with a compression-based approach to data clustering and classification  \citep{ma2007segmentation,wright2008classification}. In the past two years, they have had frequent discussions and debates about understanding (deep) learning and (artificial) intelligence.  Their  shared interests in intelligence have brought all these fundamental  ideas together and led to the recent collaboration on closed-loop transcription \citep{dai2021closedloop}, and eventually to many of the views shared in this paper. Doris is deeply interested in whether and how the brain implements generative models for visual perception, and her group has been having intense discussions with Yi on this topic since moving to UC Berkeley a year ago. 

The idea of writing this position paper is partly motivated by recent stimulating discussions among a group of researchers with very diverse backgrounds in artificial intelligence, applied mathematics, optimization, and neuroscience: Professor John Wright and Stefano Fusi of Columbia University,  Professor Yann LeCun and Rob Fergus of New York University,  Dr. Xin Tong of MSRA. We realize that these  perspectives might be interesting to broader scientific and engineering communities.

Some of the thoughts presented about integrating pieces of the puzzle for intelligent systems can be traced back to an advanced robotics course that Yi had led and organized jointly with Professor Jitendra Malik, Shankar Sastry, and Claire Tomlin as Berkeley EECS290-005: {\em the Integration of Perception, Learning, and Control} in Spring 2021. The need for an integrated view or a ``unite and build'' approach seems to be a topic that is drawing increasing interest and importance for the study of Artificial Intelligence.  

We would also like to thank many of our former and current students who, against extraordinary odds, have worked on projects under this new framework in the past several years when some of the ideas were still in their infancy and seemed not in accordance with the mainstream, including Xili Dai, Yaodong Yu, Peter Tong, Ryan Chan, Chong You, Ziyang Wu, Christina Baek, Druv Pai, Brent Yi, Michael Psenska, and others. Many of the technical evidence and figures used in this position paper are conveniently borrowed from their recent research results.

\end{document}